\documentclass[preprint,12pt,authoryear]{elsarticle}

\usepackage{hyperref}
\hypersetup{
    colorlinks=true,
    linkcolor=blue,
    filecolor=magenta,
    urlcolor=cyan,
    citecolor=blue,
}
\usepackage{multirow}
\usepackage{booktabs}
\usepackage{caption}
\usepackage{tabularx}
\usepackage{graphicx}
\usepackage{xcolor}
\usepackage{subcaption}

\usepackage{algorithm}
\usepackage{algpseudocode}
\usepackage{makecell}
\usepackage{amsmath}

\usepackage{amssymb}

\usepackage{lineno}

\begin{document}

\begin{frontmatter}



\title{Applying Unsupervised Semantic Segmentation to High-Resolution UAV Imagery for Enhanced Road Scene Parsing \tnoteref{lable1}}




\author[label1]{Zihan Ma}
\ead{zihanma@umd.edu}

\author[label2]{Yongshang Li \corref{cor1}}
\ead{yshli@chd.edu.cn}
\cortext[cor1]{Corresponding author}

\author[label2]{Ronggui Ma}
\ead{rgma@chd.edu.cn}

\author[label2]{Chen Liang}
\ead{chliang@chd.edu.cn}



\affiliation[label1]{organization={Department of Civil \& Environmental Engineering, University of Maryland},
            city={College Park},
            postcode={20742}, 
            state={MD},
            country={USA}}
            
\affiliation[label2]{organization={School of Information Engineering, Chang'an University},
            city={Xi'an},
            postcode={710064}, 
            state={Shaanxi},
            country={China}}

\begin{abstract}
There are two challenges presented in parsing road scenes from UAV images: the complexity of processing high-resolution images and the dependency on extensive manual annotations required by traditional supervised deep learning methods to train robust and accurate models. In this paper, a novel unsupervised road parsing framework that leverages advancements in vision language models with fundamental computer vision techniques is introduced to address these critical challenges. Our approach initiates with a vision language model that efficiently processes ultra-high resolution images to rapidly identify road regions of interest. Subsequent application of the vision foundation model, SAM, generates masks for these regions without requiring category information. A self-supervised learning network then processes these masked regions to extract feature representations, which are clustered using an unsupervised algorithm that assigns unique IDs to each feature cluster. The masked regions are combined with the corresponding IDs to generate initial pseudo-labels, which initiate an iterative self-training process for regular semantic segmentation. Remarkably, the proposed method achieves a mean Intersection over Union (mIoU) of 89.96\% on the development dataset without any manual annotation, demonstrating extraordinary flexibility by surpassing the limitations of human-defined categories, and autonomously acquiring knowledge of new categories from the dataset itself.
\end{abstract}









\end{frontmatter}


\section{Introduction}
\label{sec:intro}

Road infrastructure plays a pivotal role in shaping the mobility, daily lives, and economic activities of individuals. Accurate road information stands as an indispensable prerequisite for efficient infrastructure management and maintenance, impacting everything from traffic management to urban planning \citep{flah2021machine, bouraima2023integrated,HU2023120897}. Traditionally, data on road infrastructure conditions are collected through manual inspections, where teams document the status of traffic signage, road markings, traffic signals, etc by walking or driving along roadways \citep{spencer2019advances, HU2023120897, nguyen2021two}. However, the manual processes are not only labor-intensive and time-consuming but also result in error and bias due to the subjectivity of human perception. Additionally, it fails to fulfill the growing demand for large-scale data collection required by modern traffic management systems.

With advancements in image sensor technology, traffic management authorities increasingly rely on vehicle-mounted cameras for road information collection \citep{zhou2022review, HU2023120897}. This approach, equipped with high-resolution cameras and vehicle-mounted global positioning system (GPS) devices, allows for the acquisition of timely, real-time, and precise data on various road infrastructure and their respective locations. Despite its efficiency, the vehicle camera-based method has limitations. It is heavily dependent on favorable transportation and weather conditions. Furthermore, in areas inaccessible to vehicles and other large equipment, this method is often inadequate, leading to potential data collection gaps. This underscores the need for alternative solutions such as unmanned aerial vehicles (UAVs), which can operate in less accessible environments and provide comprehensive data coverage for effective road scene parsing.

Presently, researchers are turning to unmanned aerial vehicles (UAVs) to gather road information \citep{outay2020applications, astor2023unmanned, iftikhar2023target, qiu2022asff, xu2023unmanned}. 
UAVs possess the capability to offer a higher vantage point, providing an aerial perspective that enables the capture of panoramic images of roadways. This perspective encompasses a broader range of infrastructure and delivers unprecedented visual detail due to ultra-high-resolution imaging capabilities. Furthermore, UAV-based photography techniques demonstrate remarkable adaptability, successfully accessing areas that pose challenges for traditional vehicular imaging technologies. This represents a comprehensive complement and significant enhancement to the data collected by vehicle-mounted cameras.

The images generated by UAVs typically feature extremely high resolution and detail, making the processing of such images a highly challenging task \citep{xu2023unmanned, iftikhar2023target, HAMZENEJADI2023120845}. Firstly, handling and analyzing large-scale, ultra-high-resolution images demand substantial computational power and efficient algorithms to ensure timely data processing. Moreover, target recognition in UAV imagery often involves identifying small objects, further escalating the complexity of the task. These small targets occupy very minimal pixel areas in the images, thus are easily susceptible to noise and environmental interferences. The advent of deep learning (DL) provide powerful tools to address these challenges.

\begin{figure}[t]
    \centering
    \includegraphics[width=\textwidth]{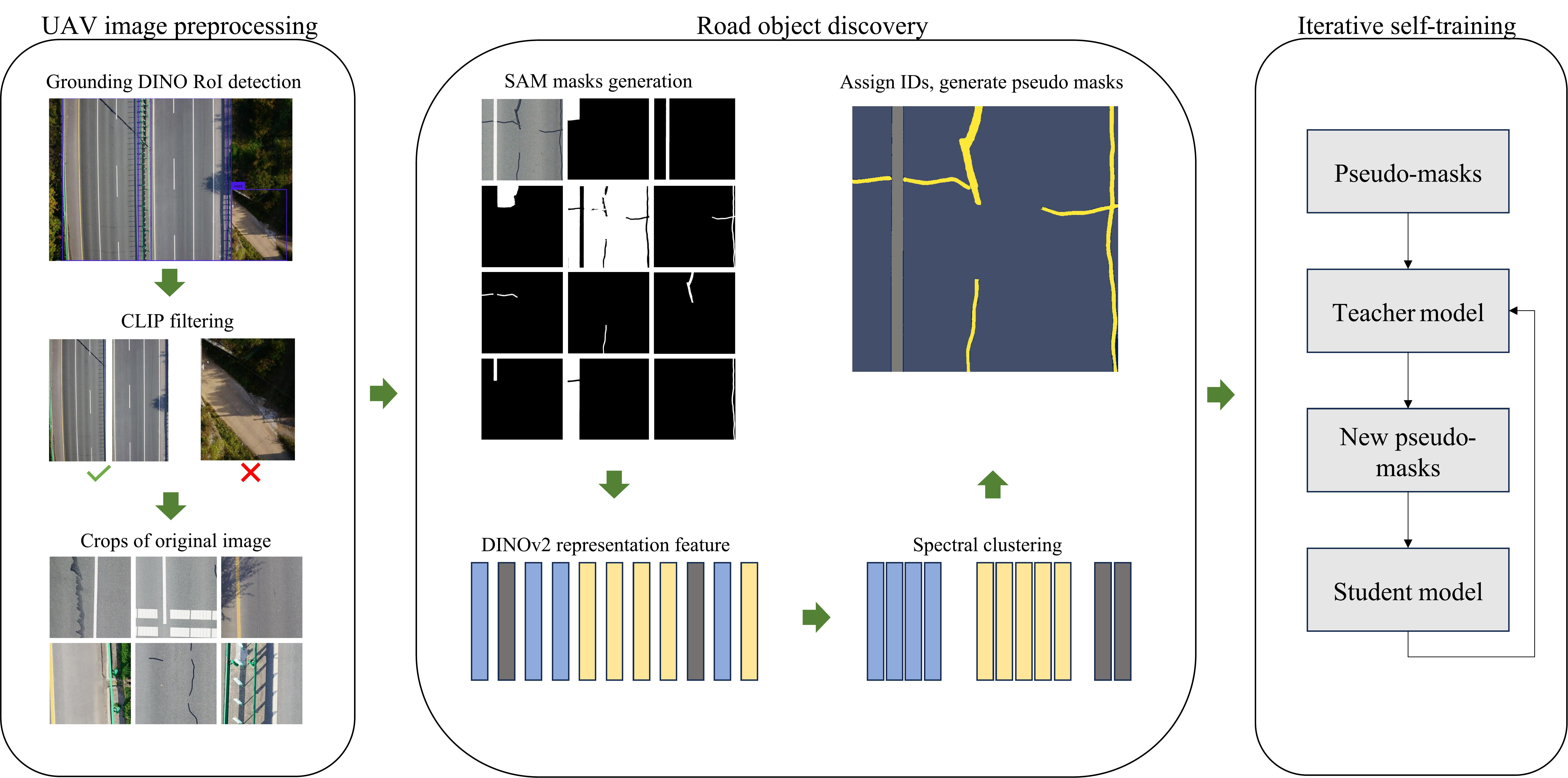}
    
    \caption{Overview of unsupervised UAV road scene parsing.}
    
    \label{fig1}
\end{figure}

Recent studies have increasingly combined UAV aerial photography with DL techniques \citep{qiu2022asff, zhang2022road, silva2023automated, senthilnath2020deep, byun2021road, zhu2022monitoring, cao2023segmentation,gao2023pixel,HU2023120897, HAMZENEJADI2023120845, JEONG2022116791, wu2023uav,chu2024cascade} such as convolutional neural networks (CNNs)\citep{krizhevsky2012imagenet} and vision transformers (ViTs)\citep{dosovitskiy2020image} to identify regions of interest (RoI) in road images. Despite these advances, most existing methods remain rooted in supervised learning paradigms, necessitating large-scale manual annotations to train accurate and robust models. The crowd-labeled process consumes a lot of time and resources, thus limiting the scalability and applicability of the DL approaches.
In response to the limitations, this paper introduces an innovative unsupervised semantic segmentation method aimed at parsing road scenes from UAV images and identifying regions of interest. In contrast to traditional supervised semantic segmentation methods, our approach eliminates the need for manually defined annotations for training, thus avoiding the problems of labeling cost and data scarcity. Simultaneously, our approach leverages the inherent flexibility of the unsupervised paradigm, allowing for the identification of diverse objects within the dataset without the constraints of predefined categories, thus facilitating open-vocabulary semantic segmentation. The framework for our proposed unsupervised road semantic parsing method is outlined in Figure ~\ref{fig1}  and comprises the following innovative components:

\begin{itemize}
\item	Preprocessing with Vision-Language Models (VLMs): The emerging VLMs are utilized for preprocessing UAV data, where the main challenge lies in efficiently processing high-resolution images and detecting RoI. An approach is proposed, employing the text-guided object detection capability of the multi-modal model Grounding DINO \citep{liu2023grounding}, to extract RoI from UAV images. To mitigate the potential issue of false detection, CLIP \citep{radford2021learning} is introduced for filtering purposes;

\item	Mask Generation with SAM: Subsequently, the vision foundation model SAM (Segment Anything Model) \citep{kirillov2023segment}, known for its potent zero-shot inference capability, is employed to generate masks for the images. However, it's worth noting that SAM currently lacks support for generating category information;

\item	Feature Extraction and Pseudo-Label Synthesis: Representation learning models \citep{he2016deep, oquab2023dinov2} are utilized to extract features corresponding to image regions related to the masks. These features will then be employed for the synthesis of pseudo-labels;

\item   Unsupervised Clustering for Category Discovery: An unsupervised clustering algorithm processes the representation features of masks, categorizing different objects and assigning a unique ID to each cluster. This step combines masks with the respective IDs to generate the supervised signals, i.e., pseudo-labels, required for training the semantic segmentation model;

\item   Iterative Self-Training: Lastly, building upon these initial pseudo-labels, we initiate an iterative self-training process \citep{xie2020self} for a regular semantic segmentation network, enhancing its learning efficiency and accuracy.
\end{itemize}

In this study, We curate a dataset comprising 22,338 UAV images, which we refer to as the drone-view road image dataset (DRID22k). Our proposed framework, denoted as clustering object masks for road parsing (COMRP), with each module undergoing rigorous experimental validation. Our contributions are summarized as follows:

\begin{enumerate}
\item	We delve into the application of emerging VLMs and vision foundation models, assessing their utility in processing UAV imagery;
\item   We explore and propose a novel unsupervised road semantic parsing framework that does not use any manual annotations as supervised signal throughout the process. Instead, it utilized unsupervised clustering to synthesize pseudo-labels and initiates an iterative self-training process, enhancing the model's learning capability;
\item	We conduct exhaustive comparative experiments to quantify the significance of different parameters within each module of the COMRP framework;
\item   To foster further research and development in unsupervised semantic segmentation within the remote sensing field, we provide open access to the DRID22k dataset \footnote{https://github.com/CHDyshli/unsupervised-road-parsing}.
\end{enumerate}

The remainder of this paper is structured as follows: Section \ref{sec:relatework} reviews the related research, providing context and background for our work. Section \ref{sec:methodology} describes our proposed methodology in detail. The detailed data processing procedure, experiments, as well as the results, are presented in Section \ref{sec:experiments}. Subsequently, we discuss the potential and limitations of our method. Finally, we summarize the contributions of this paper and outline prospects for future work.


\section{Related Work}
\label{sec:relatework}
Two fields are relevant to the research in this paper, including UAV-based road scene parsing, and unsupervised semantic segmentation.

\subsection{UAV-based Road Scene Parsing}

The utilization of unmanned aerial vehicles (UAVs) for road scene parsing has gained traction due to their cost-effectiveness and operational flexibility. This emerging interest has spurred the development of advanced UAV image processing algorithms, especially those incorporating deep neural networks \citep{qiu2022asff, zhang2022road, silva2023automated, senthilnath2020deep, byun2021road, zhu2022monitoring, cao2023segmentation, gao2023pixel}. 

The first line of work aims to detect objects of interest, determine their coordinates, and classify them in UAV imagery, a task commonly referred to as object detection. This type of research often faces a challenge when dealing with objects in images with very small pixel areas. In the realm of detecting and recognizing traffic elements, \citet{qiu2022asff} proposed an adaptive spatial feature enhancement method for YOLOv5, addressing issues related to dense elements and poor multi-scale object detection performance in UAV imagery, particularly in the detection of small objects. 
On the other hand, \citet{zhang2022road} found that the accuracy of road damage detection in UAV imagery using YOLOv3 was low. Their analysis revealed insufficient communication between the backbone network and feature fusion module. To enhance overall model performance, they introduced a multi-level attention mechanism between these components.
\citet{silva2023automated} conducted a more comprehensive comparison, evaluating YOLOv4, YOLOv5, and YOLOv7 algorithms for detecting and locating road damage in UAV imagery. In the field of traffic surveillance, \citet{zhu2022monitoring} has established an AI and UAV-based platform for monitoring construction sites on roads. This platform employs YOLOv4 for safety factor detection and utilizes DeepSORT for target tracking. 
Alternatively, \citet{byun2021road} proposed the utilization of deep learning to analyze and detect vehicles in UAV-recorded videos. They also incorporated distance measurements between lanes to calculate image proportions, enabling further estimation of vehicle speeds. This approach culminated in the development of a comprehensive framework for road traffic monitoring.

Another line of work focuses on a more fine-grained detection approach, where researchers aim to classify each pixel in the image, a task known as semantic segmentation. \citet{senthilnath2020deep} attempted to extract road from images captured by drones. The method employed involved acquiring knowledge from publicly available datasets and transferring the trained full convolution network to different application scenarios.
On a more challenging note, \citet{cao2023segmentation} endeavored to detect road cracks in UAV imagery. The authors integrated SENet attention mechanisms and blurpool pooling into UNet, thereby addressing issues related to cracks being influenced by background noise and the segmentation results appearing discontinuous. Similarly, \citet{gao2023pixel} also employed an improved UNet for segmenting road cracks in remote sensing images.

In the aforementioned literature, whether the authors focused on object detection or semantic segmentation, they remained within the paradigm of supervised learning. The substantial demand for human-defined annotations continues to be a bottleneck for these methods.

\subsection{Unsupervised Semantic Segmentation}
\label{subsec_2.2}

Unsupervised semantic segmentation of remote sensing images is a relatively uncharted territory. However, in typical natural scenes like COCO \citep{lin2014microsoft} and PASCAL VOC \citep{pascal-voc-2012}, there has been some pioneering work in unsupervised semantic segmentation. Learning semantic representations directly from pixels without any available manual annotations is quite challenging, leading many unsupervised methods to rely on additional prior knowledge.

\citet{van2021unsupervised} introduced the MaskContrast, which learns pixel embeddings through contrasting salient objects. This approach primarily utilizes an unsupervised saliency estimator to generate mask proposals in the images, which include intermediate image features such as boundaries and shapes. These masks, serving as priors, initiate a contrastive learning framework for dense pixel features.

\citet{cho2021picie} proposed PiCIE, which leverages clustering results of multiple regions in images as priors. Specifically, each image is divided into several regions, with the representation features for each region representing low-level characteristics like color, texture, and edges. This can serve as a coarse division of semantic concepts.

\citet{melas2022deep} discovered that they could utilize the Laplacian eigenvectors of the feature affinity matrix from the self-supervised network (e.g., DINO \citep{caron2021emerging}) to achieve image segmentation and object localization. These feature vectors were already capable of decomposing the image into meaningful parts. By clustering these segmented features on the dataset, they were also able to obtain semantic segmentation. Similarly, \citet{hamilton2022unsupervised} also harnessed the capabilities of the unsupervised feature learning framework, DINO, which can generate dense features with consistent semantic relevance.
The authors of COMUS \citep{zadaianchuk2022unsupervised} combined various aspects of previous work. They employed an unsupervised saliency detector, DeepUSPS, to estimate proposal masks for images. Subsequently, DINO served as a representation learning network to extract feature vectors from these masks.

Our work is an extension of these studies; however, we primarily focus on the more challenging task of unsupervised semantic segmentation of UAV remote sensing images.


\section{Methodology}
\label{sec:methodology}
As illustrated in Figure \ref{fig1}, our approach is structured into three primary components, each designed to address specific challenges in the processing pipeline: preprocessing of UAV images, road object discovery, and iterative self-training.

\subsection{UAV Image Preprocessing}
\label{subsec:uavImagePreprocessing}
We perform the preprocessing for two reasons. First, the high resolution of the UAV images makes the processing of the raw-resolution images time-consuming. Second, our study focuses on road objects, while a significant portion of the content in the images pertains to the surrounding environment (e.g., vegetation). 
Therefore, we introduce the most recent VLMs to address these challenges. VLMs are trained using web-scale supervised signals of plain text, such as image captions, subject labels, and descriptions, which are almost infinitely accessible on the Internet \citep{zhang2023vision}. This alternative to traditional fixed-category and crowd-labelled supervised signals significantly enhances the performance and expressiveness of the image encoder. 
This feature makes VLMs highly suitable for a variety of vision downstream tasks, such as open-set object detection \citep{liu2023grounding} and zero-shot image classification \citep{radford2021learning}.

\begin{figure}[t]
    \centering
    \includegraphics[width=\textwidth]{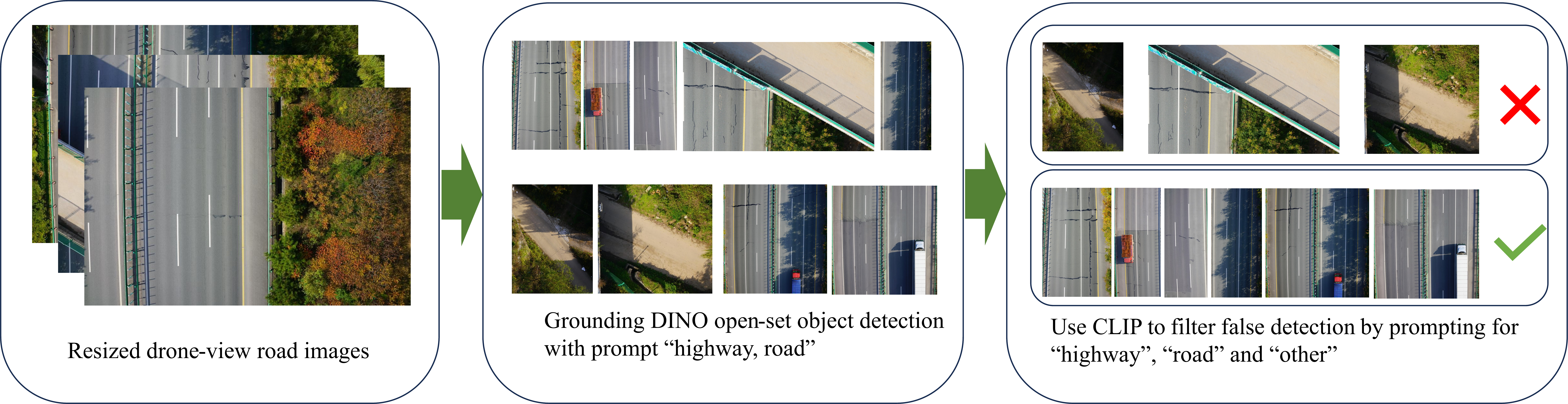}
    
    \caption{Vision language model-based UAV image preprocessing.}
    
    \label{fig2}
\end{figure}

As shown in Figure \ref{fig2}, our proposed preprocessing method consists of two steps: RoI detection and false detection filtering. Firstly, we employ Grounding DINO \citep{liu2023grounding} to detect road regions in the resized UAV images, thereby eliminating environmental influences. Grounding DINO is a vision transformer-based \citep{dosovitskiy2020image} open-set object detection model capable of detecting objects of any category in images based on text input. Its key capability lies in zero-shot inference without requiring any fine-tuning, even image data is out-of-distribution, making it highly suitable for our application. In this step, we set the input for the text encoder as "road, highway." Grounding DINO can detect the road areas of interest, but it may also produce some false detections, such as "side roads" or "overpasses on the road" (see Figure \ref{fig2} in the middle). To address this issue, we introduce the second step: employing CLIP \citep{radford2021learning} for filtering out undesired detections. 
The CLIP model, during its pre-training phase, employs a vast corpus composed of 400 million pairs of image-text for contrastive learning. It utilizes a symmetric cross-entropy loss function, aiming to maximize the cosine similarity between matching image-text pairs while simultaneously minimizing the similarity between non-matching image-text pairs. CLIP's model design makes it particularly suitable for zero-shot image classification by calculating the similarity between images and text. Specifically, we input candidate regions from the first step into CLIP's image encoder, and on the text side, we employ three prompts: "road," "highway," and "other." When a candidate region best matches the prompt "other," we filter it out, retaining the images representing our desired road region (see Figure \ref{fig2} in the right).

\subsection{Road Object Discovery}
In Algorithm \ref{alg:road_object}, we present the steps for road object discovery.
\begin{algorithm}[h]
\caption{Road Object Discovery for Pseudo-label Generation}\label{alg:road_object}
\begin{algorithmic}
\State \textbf{Given:} $ N $ road images $x_i$, mask generator $M$, mask area threshold $\theta$ (pixel), representation learning network $R$, clustering method $C$. 
\State \textbf{Step 1:} Generate all proposal mask $m_i$ for road image $x_i$ by $M(x_i) > \theta $. Obtain object region $o_i$ based on $m_i$,  $o_i=bounding\_box(x_i, m_i)$.
\State \textbf{Step 2:} Compute representation feature vector $r_i$ of road object region $o_i$, $r_i=R(o_i)$.
\State \textbf{Step 3:} Cluster representation feature vector $r_i$ by spectral clustering method $C$. Assign cluster ID $d_i$ to each object region $o_i$.
\State \textbf{Step 4:} Combine cluster ID $d_i$ and proposal mask $m_i$ to get initial pseudo-label $l_i$.
\State \textbf{Return:} Road image pseudo-label $l_i$.
\end{algorithmic}
\end{algorithm}

\subsubsection{Segment Anything on the Road}
As described in Section \ref{subsec_2.2},  many researchers often employ prior knowledge or assumptions to simplify unsupervised semantic segmentation. 
The prior knowledge used in the COMRP is derived from SAM\citep{kirillov2023segment}, which is a generalized segmentation model capable of generating masks for any object in any image. 
The core capability of SAM is reflected in its zero-shot inference, which means that it is able to perform effective image segmentation without previous domain-specific training samples. However, it's worth noting that the masks currently generated by SAM lack category information (Although the authors have conducted image segmentation based on text prompts, the related code and model are not publicly disclosed). SAM supports prompt-based segmentation and automatic segmentation. 
Manual prompt-based image segmentation relies on foreground and background points or object bounding boxes. Providing segmentation prompts manually is a daunting task for large-scale mask generation and deviates from our unsupervised framework.

Building upon above analysis, we employ SAM for automated mask generation on the DRID22k, as detailed in \ref{subsec:maskgeneration}. We further investigate the impact of various SAM parameters  on the results of road object mask generation.

\subsubsection{Representation Learning for Mask Region}
In machine learning, the key role of representation learning is that it enables the model to autonomously extract hierarchical and meaningful feature representations from raw data\citep{bengio2013representation}. The learned representation features better reflect the underlying structure and patterns within the data. Specifically in computer vision, a model's intermediate representations abstract the content and structure of images at different levels \citep{krizhevsky2012imagenet}. In this paper, we employ the CNN-based ResNet50 \citep{he2016deep} and ViT-based DINOv2 \citep{dosovitskiy2020image,oquab2023dinov2}, as the models for representation learning.

The ResNet \footnote{https://huggingface.co/microsoft/resnet-50} we use is pretrained in a supervised manner on ImageNet-1k \citep{deng2009imagenet}. DINOv2 \citep{oquab2023dinov2}, on the other hand, extends the DINO \citep{caron2021emerging} by scaling up both the model and the dataset, employing self-supervised pre-training. To reduce computational cost, we opt for the smaller model ViT-B trained by DINOv2 \footnote{https://huggingface.co/facebook/dinov2-base}.

We extract the representation feature for each object region using the pre-trained ResNet and ViT-B. These regions are cropped from the original images based on the bounding boxes of the masks and resized to 224 × 224. Features from different layers of the model represent abstractions at various levels of the input image. Most research typically utilizes the final output of the model as the representation for the entire image, such as the 7 × 7 × 2048 feature representation of ResNet or the last layer's CLS token of the ViT. These features are commonly used as the basis for classification tasks. However, a question arises when the model's input is only a part of the image, as in our case with the mask regions: can the output from the final layer still effectively represent these regions? To explore this issue, we conduct detailed comparative experiments in Section \ref{subsec:representation}.

\subsubsection{Clustering Masks for Pseudo-labels}
We employ the spectral clustering \citep{von2007tutorial} method to cluster the mask feature vectors generated by the representation network. We observe that, owing to the capacity of the representation model, different masks are clustered together based on attributes such as color, shape, and size. Here arises an issue: when SAM generates masks for an input image, a single object is often segmented into distinct parts. For instance, a guardrail may be divided into an upper railing and lower support pillar. In this case, we perform cluster merging, where sub-clusters sharing the same category are merged into  a new unified cluster. Subsequently, we assign IDs to the newly merged clusters. At this point, all labeled masks for an image constitute the pseudo-label for that image.

During clustering of the SAM-generated masks, we note that not all clusters contain only one category, and therefore it is not possible to specify unique cluster ID. Therefore, these clusters will be discarded.

In addition to spectral clustering, we also compare the results with the k-means, k-medoids, and agglomerative clustering methods. For specific results, please refer to Section \ref{subsec:clusteringmasksforpseudolabels}.

\subsection{Iterative Self-training}

Self-training is a semi-supervised learning method that leverages labeled data to train a teacher model and infers pseudo-labels on unlabeled data. Subsequently, both the labeled and pseudo-labeled data are used to train a student model \citep{xie2020self}. \citet{zhu2021improving} and \citet{du2022learning} have already discovered that self-training can effectively enhance model performance in semantic segmentation tasks.

Our approach is more radical because we do not use any annotated images as training data at all. With the pseudo-labels generated from the clustering method in the COMRP, we can train any regular semantic segmentation network as the teacher model. In this paper, we explore two different types of networks, the CNN-based Deeplabv3p \citep{chen2018encoder} and the ViT-based SegFormer \citep{xie2021segformer}. Once the teacher model's training is complete, it can generate its own pseudo-labels. On these new pseudo-labels, a student model, structured similarly to the teacher model, is used to perform iterative self-training until performance no longer improves.

\section{Experiments and Results}
\label{sec:experiments}

\subsection{Dataset}

\subsubsection{Data Collection}

\begin{figure}[t]
    \centering
    \includegraphics[width=\textwidth]{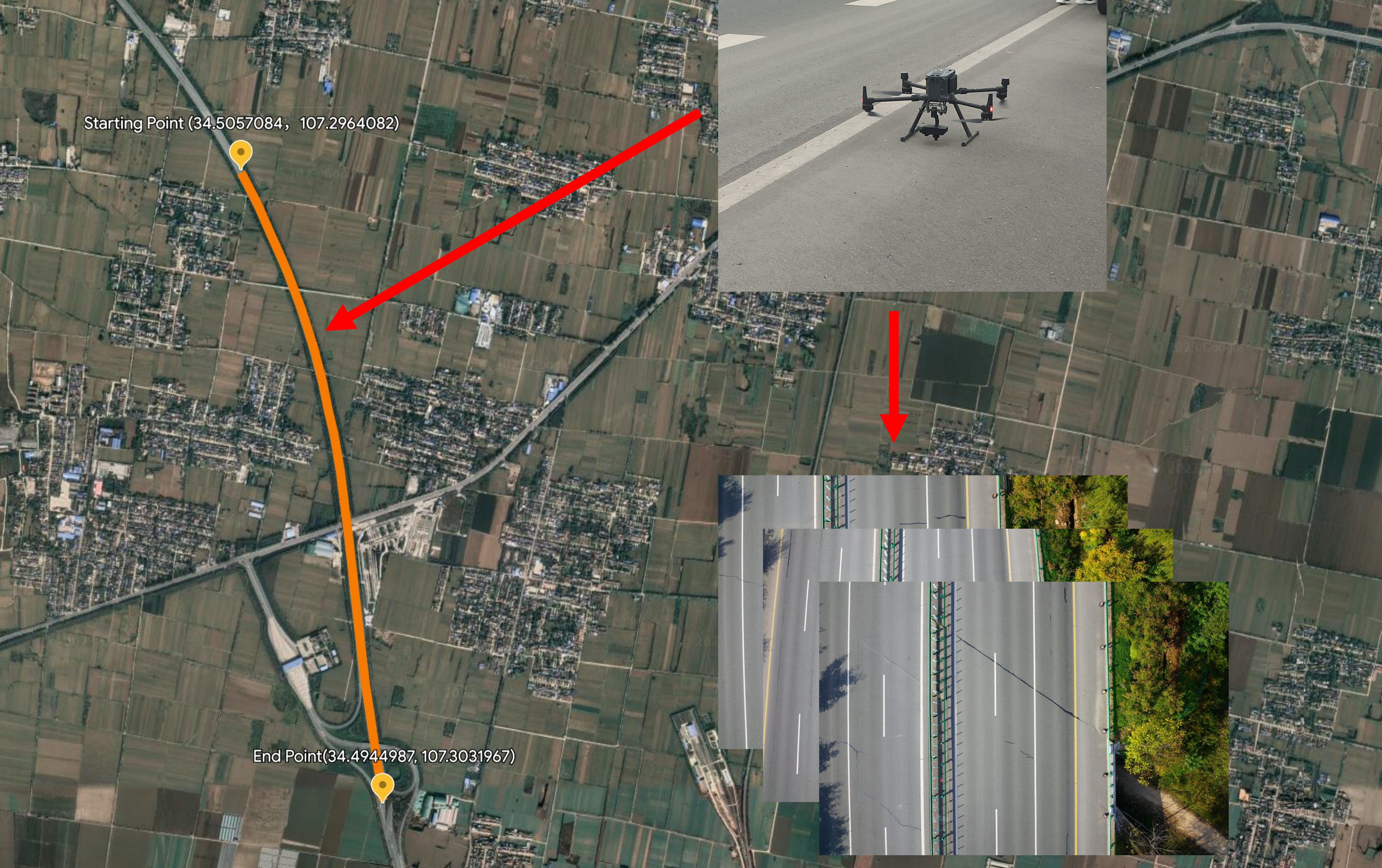}
    \caption{Data collection.}
    \label{fig3}
\end{figure}

The UAV image was collected from the Yinkun Expressway within the Shaanxi province of China, as illustrated in Figure \ref{fig3}. The geographical coordinates range from the starting point (34.5057084, 107.2964082) to the end point (34.4944987, 107.3031967). A DJI Zenmuse P1 drone was used for the aerial photography, and the specific parameters of the camera are provided in Table \ref{tab1}. To ensure the clarity of road objects, we set the flight altitude throughout the journey at 40 meters, with an image resolution of 7952 × 5304 and a spatial resolution of 0.25 cm/pixel. We collected a total of 447 images, and the subsequent section will provide a detailed description of the data processing steps.

\begin{table}[t]
    \centering
    \begin{tabular}{cc}
    \toprule
        Parameters	& Value	\\
        \midrule
        Aperture		& f/5.6			\\
        Exposure		& 1/1000 sec.	\\
        Focal Length    & 	35.0 mm      \\
        ISO             & 	250          \\
        Metering        & Center-weighted average\\
        \bottomrule
        \end{tabular}
    \caption{UAV camera parameters.}
    \label{tab1}
\end{table}

\subsubsection{UAV Road Image Preprocessing}

As described in Section \ref{subsec:uavImagePreprocessing} , the primary objective of preprocessing UAV imagery is to address the significant computational demands posed by high-resolution images and to detect regions of interest.
Our proposed approach comprises two main steps: detecting road areas in resized images and removing false positives. In the first step, the longer side of the original high-resolution image is resized to 1024 pixels, maintaining the aspect ratio. The resized image is then fed into the image encoder of Grounding DINO \citep{liu2023grounding}, where we utilize the smallest swin-transformer tiny \citep{liu2021swin} to reduce computational load. The text encoder defaults to BERT-base \citep{devlin2018bert}. We experimented with two text prompts, ``road" and ``highway," and observed no significant differences in the results. In this step, the two crucial hyperparameters of Grounding DINO, namely ``box\_threshold" and ``text\_threshold", are set to 0.35 and 0.25, respectively. Figure \ref{fig2} illustrates the detection results for the road regions, demonstrating accurate detection alongside some false positives. Although these false-positive regions (see \ref{fig2} in the center) could literally be considered ``road", they are not the regions of interest to this study.

To address the issue in the first step, we propose using CLIP \citep{radford2021learning} for filtering out the false detections. The distinctive feature of the CLIP is its ability to perform zero-shot image classification by measuring the similarity score between an image and text. For the CLIP text encoder, we utilize three text inputs, i.e., ``road", ``highway" and ``other",  while the input for the image encoder is the road regions detected by Grounding DINO. The similarity score between an image and text is converted into a probability using the softmax function. When the probability of matching with ``other" is the highest, we consider that the region is not the ``road" of interest. 
After these two steps, we have essentially completed the detection of road regions.

It's worth noting that all the previous image processing was conducted on the resized images. However, resizing high-resolution UAV images results in a significant reduction in quality, which is detrimental to fine-grained, dense prediction tasks. 
Based on the road detection results, we crop the road area from the original resolution UAV images, which ensures that small objects on the road will not be blurred due to image resizing.  Next, we crop the road region into sub-images with a resolution of 800 × 800 pixels, forming a collection of 22,338 images that served as the foundation for the subsequent study. We entitle this dataset as drone-view road image dataset (DRID22k).

\begin{figure}[t]
    \centering
    \includegraphics[width=\textwidth]{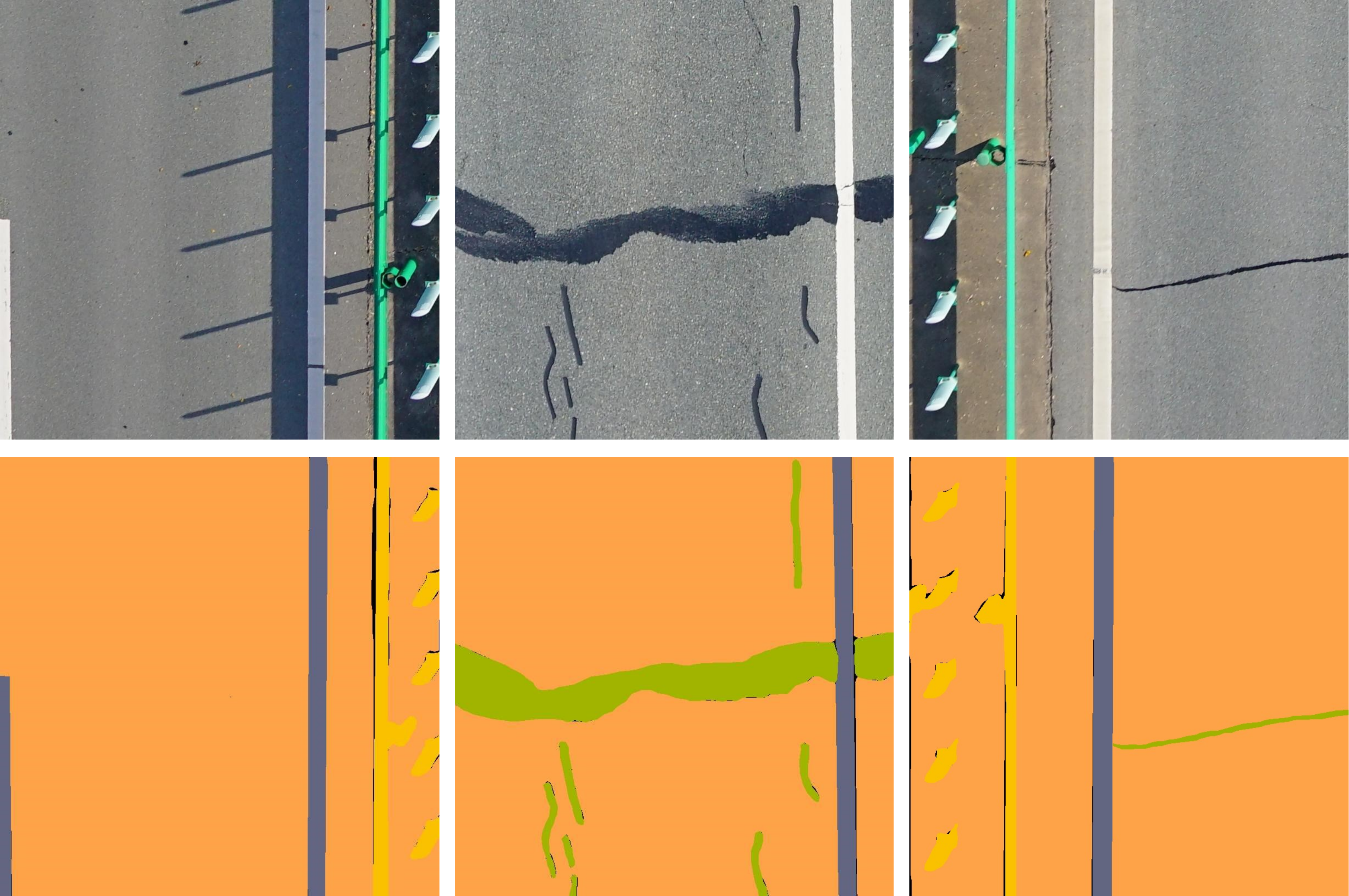}
    \caption{Sample images and corresponding annotations in DRID22k-dev.}
    \label{fig4}
\end{figure}

\subsubsection{Development Set}
Even though our objective is to perform road semantic segmentation without any supervised signal from human, we still require manually annotated image labels as ground truth to evaluate the method's performance. Following Cityscapes \citep{cordts2016cityscapes}, we perform pixel-level annotations for all the predominant object categories appearing on the road, including road, white marking, yellow marking, guardrail, crack sealing and traffic cone. To expedite our experimental progress, we employ the interactive annotation tool, EISeg \citep{hao2022eiseg}, and opt for the lightweight model HRNet18\_OCR48 \footnote{https://github.com/PaddlePaddle/PaddleSeg/tree/release/2.9/EISeg}. We randomly selected 500 images from DRID22k to serve as the development set for the entire method, hereinafter referred to as DRID22k-dev for convenience. The sample images and corresponding annotations are depicted in Figure \ref{fig4}.

\subsection{Mask Generation}
\label{subsec:maskgeneration}

Previous unsupervised semantic segmentation methods \citep{van2021unsupervised, zadaianchuk2022unsupervised} relied on saliency detectors to predict object region masks. However, these methods are greatly limited as they focus only on the main objects in an image, posing challenges in achieving fine-grained panoptic segmentation. SAM \citep{kirillov2023segment}, as a universally trained segmentation foundation model on 11 million high-quality images, can provide fairly precise object masks, effectively addressing the aforementioned challenge. In the following, SAM is employed as the mask generator.

\begin{figure}[t]
    \centering
    \includegraphics[width=\textwidth]{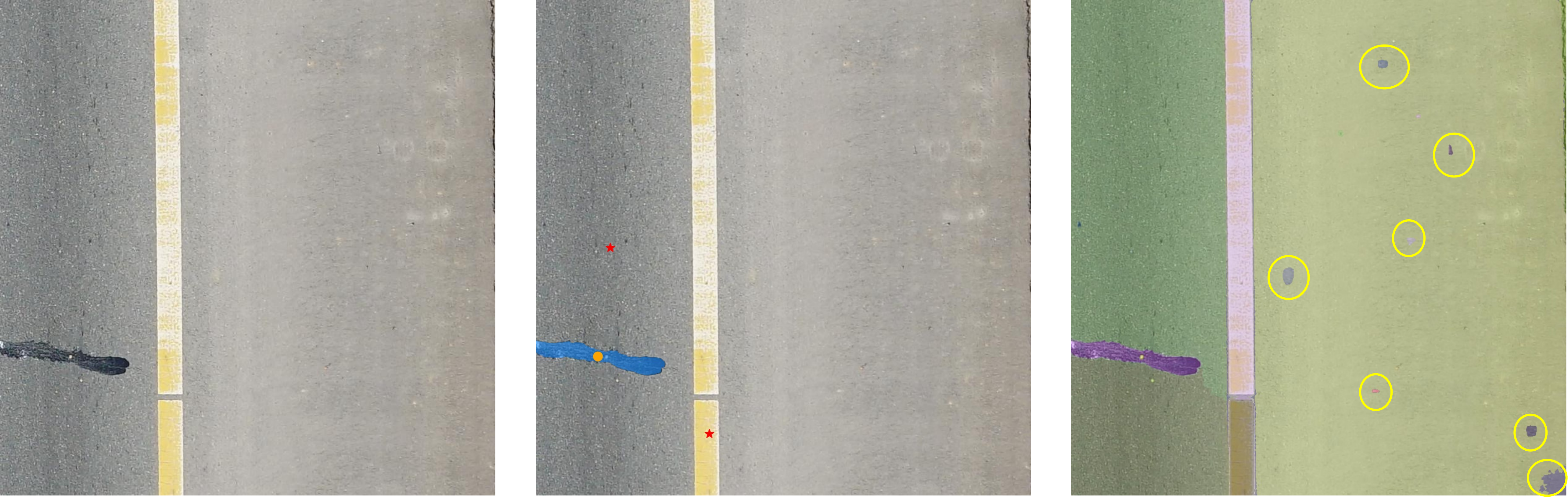}
    \caption{\textbf{(a)} Original image. \textbf{(b)} SAM prediction with one foreground point (orange) and two background points (red). \textbf{(c)} Automatic mask generation with 64×64  grid points.}
    \label{fig5}
\end{figure}

SAM aspires to enable semantic segmentation to perform zero-shot or few-shot learning akin to most recent advances in natural language processing, relying on prompting techniques \citep{brown2020language}. This entails the model segmenting regions of interest based on human prompts (foreground or background points, bounding boxes, rough masks, plain text). As exemplified in Figure \ref{fig5}b, the segmentation of crack sealing necessitates the manual delineation of foreground and background points.
To create an extremely large dataset with accurate masks, SAM's creators introduce an automatic mask generation paradigm. In the automatic mode, prompts are not precisely specified manually, but rather follow a regular grid pattern, such as 16×16, 32×32, or 64×64. Figure \ref{fig5}c illustrates the segmentation results with regular grid points. Although the manual mode can accurately segment regions of interest by providing precise foreground and background points, it is highly inefficient. The workload of generating large-scale masks in practical applications is tantamount to manual annotation. Grid-based methods can automatically generate all masks, but a large number of non-existent objects are erroneously imagined, as illustrated in the Figure \ref{fig5}c. To maintain the unsupervised paradigm, we opt for the automatic mask generation manner.

\begin{figure}[t]
    \centering
    \includegraphics[width=\textwidth]{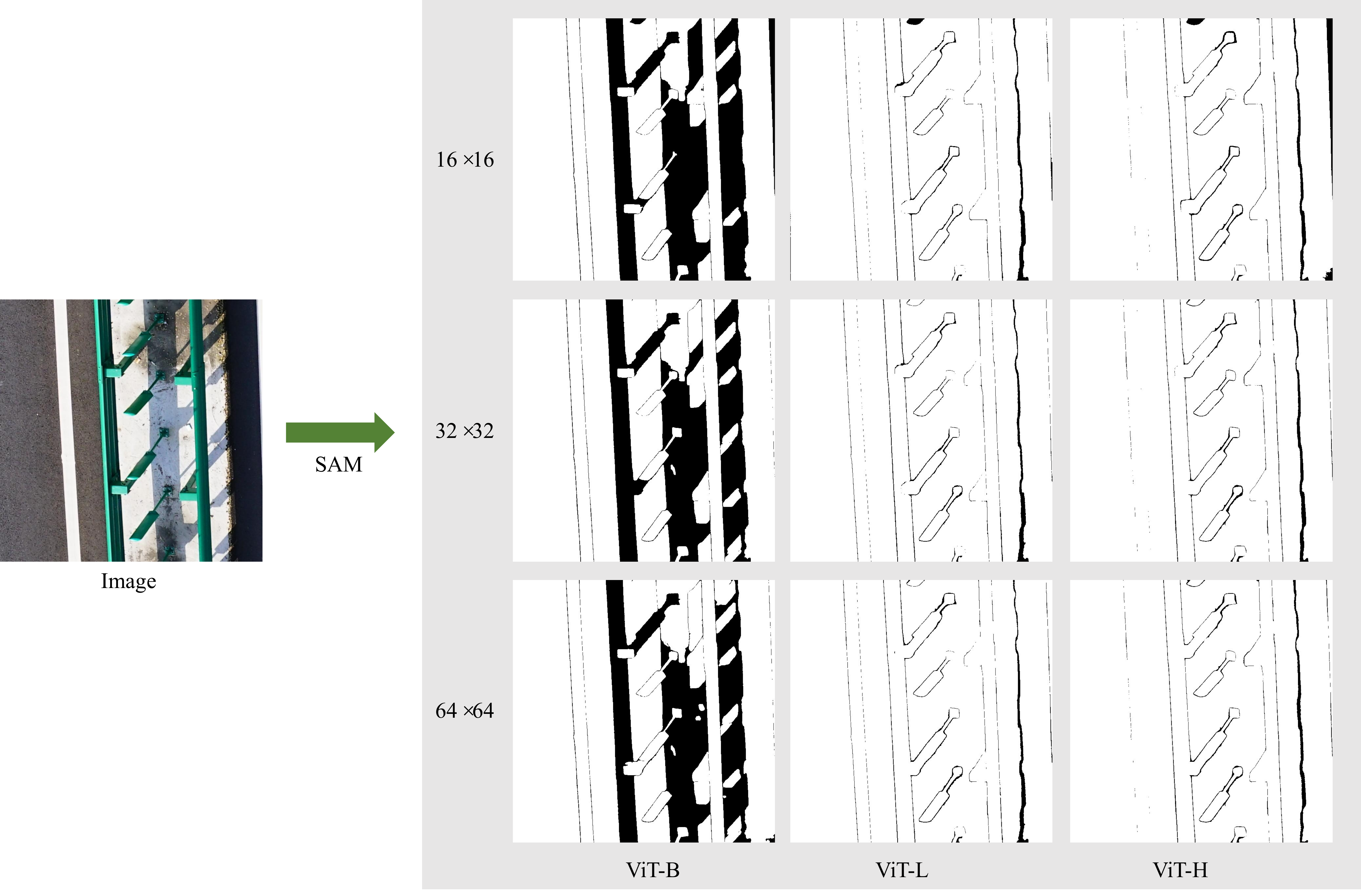}
    \caption{Impact of different image encoders and grid patterns on mask coverage. }
    \label{fig6}
\end{figure}

The density of grid points and the size of the image encoder in SAM have a substantial impact on the quality and quantity of generated masks. To investigate and quantify the significance of grid point density and model size, we investigated three different scales of image encoders (ViT-B, ViT-L, ViT-H) combined with three regular grid patterns (16×16, 32×32, 64×64). Table \ref{tab2} illustrates the mask coverage, the inference time per single image, and the quantity of generated masks on the DRID22k-dev. The mask coverage denotes the degree of overlap between the masks generated by SAM and the original image. Its magnitude signifies the richness of the masks predicted by SAM, with higher coverage implying more objects being segmented. Figure \ref{fig6} illustrates the impact of different image encoders and grid patterns on mask coverage. It is evident that ViT-B consistently achieves the lowest coverage, while ViT-L and ViT-H show no significant difference.


\begin{table}[t] 
\resizebox{\linewidth}{!}{
\begin{tabular}{cccccccccc}
\toprule
\multirow{2}{*}{Grid}    & \multicolumn{3}{c}{ViT-B} & \multicolumn{3}{c}{ViT-L} & \multicolumn{3}{c}{ViT-H} \\
    & Coverage & Time & Masks  & Coverage & Time & Masks  & Coverage & Time & Masks \\
\midrule
16×16 & 92.60 & 1.074 & 4440 & 97.85 & 1.606 & 5613 & 97.99 & 2.074 & 5426 \\
32×32 & 94.57 & 3.428 & 7442 & 98.56 & 4.312 & 9357 & 98.65 & 4.494 & 9016 \\
64×64 & 94.55 & 12.73 & 11989 & 98.58 & 15.42&14235& 98.78&16.654 &13760 \\
\bottomrule
\end{tabular}
}
\caption{Impact of model size and grid point density on mask generation.}
\label{tab2}
\end{table}

In Table \ref{tab2}, we observed that different grid point densities resulted in significant variations in the number of generated masks. Under the same model, the total number of masks generated by SAM doubles with an increase in grid point density. For instance, ViT-B with a 16×16 grid generates 8.88 masks per image, while a 64×64 grid produces 23.99 masks per image. The latter yields an average number of masks that far exceeds the reasonable number of objects present in the image, indicating that the model hallucinates many non-existent objects, as shown in Figure \ref{fig5}c.

\begin{figure}[t]
    \centering
    \includegraphics[width=\textwidth]{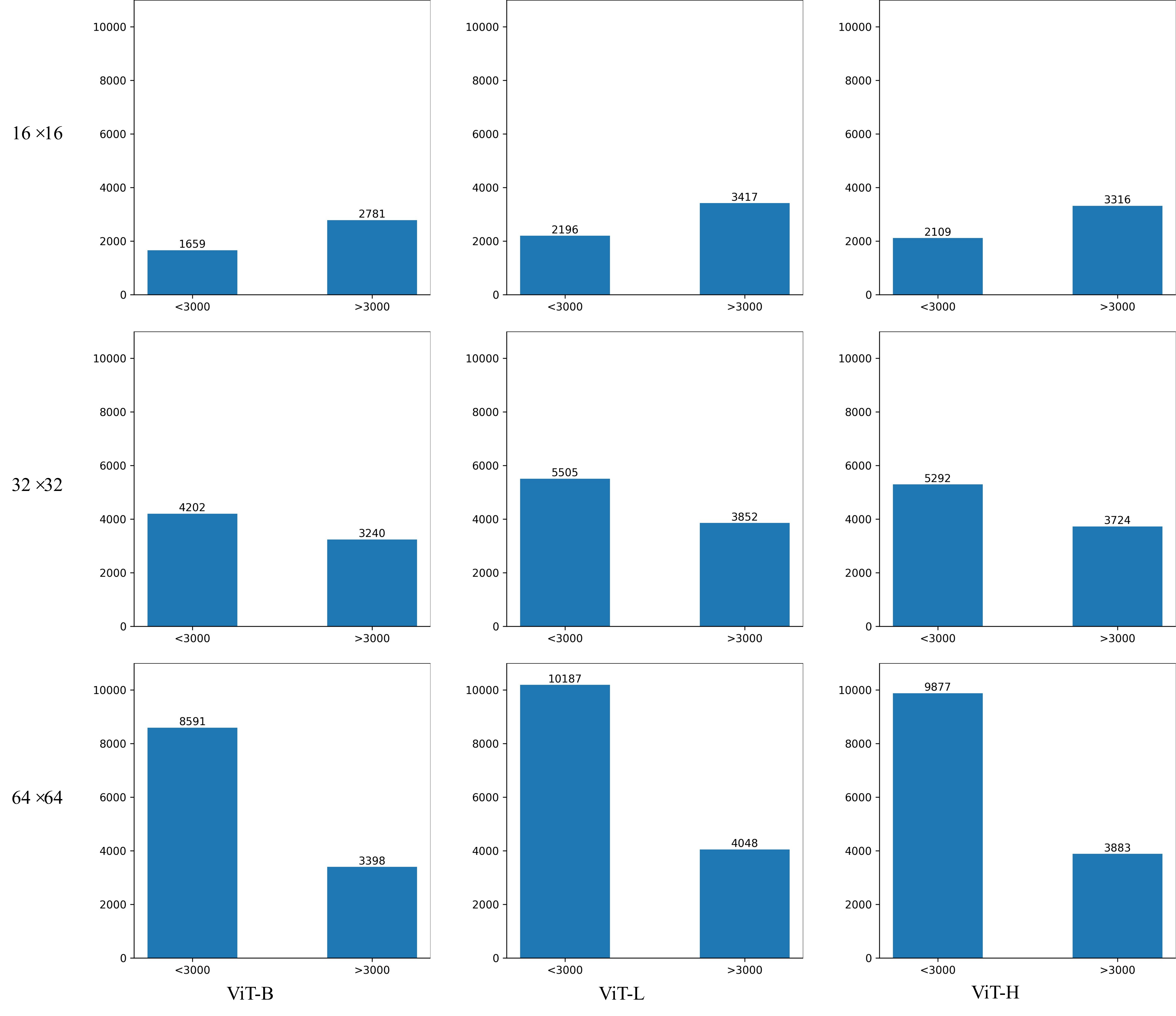}
    \caption{Number of masks with an area less than 3000 and greater than 3000 under different combinations of grids and models. }
    \label{fig7}
\end{figure}

Given that insignificant small-area masks impose a burden on subsequent feature representation learning and affect the performance of clustering algorithms, we counted the number of masks with areas smaller than and greater than 3000, considering different combinations of grids and models, as shown in Figure \ref{fig7}. We observed that with an increase in grid point density, the number of small-area masks (less than 3000) increases sharply, while the count of relatively larger masks (greater than 3000) remains stable. This indicates that an increase in grid point density leads to a substantial rise in the number of meaningless phantom masks. Notably, the area threshold of 3000 here represents a hyperparameter. In preliminary experiments, we noted that most meaningless masks have an area smaller than 3000.

Considering the observation above, in our subsequent experiments, we employ a setting of ViT-L and a 16×16 grid point, ensuring a balance between efficient inference speed and mask generation quality. We discard ViT-B due to its lower coverage, with 7.4\% of image regions not designated as any object. The exclusion of ViT-H is rooted in its substantial computational demands, coupled with its performance closely mirroring that of ViT-L.

\begin{figure}[t]
    \centering
    
    \includegraphics[width=\textwidth]{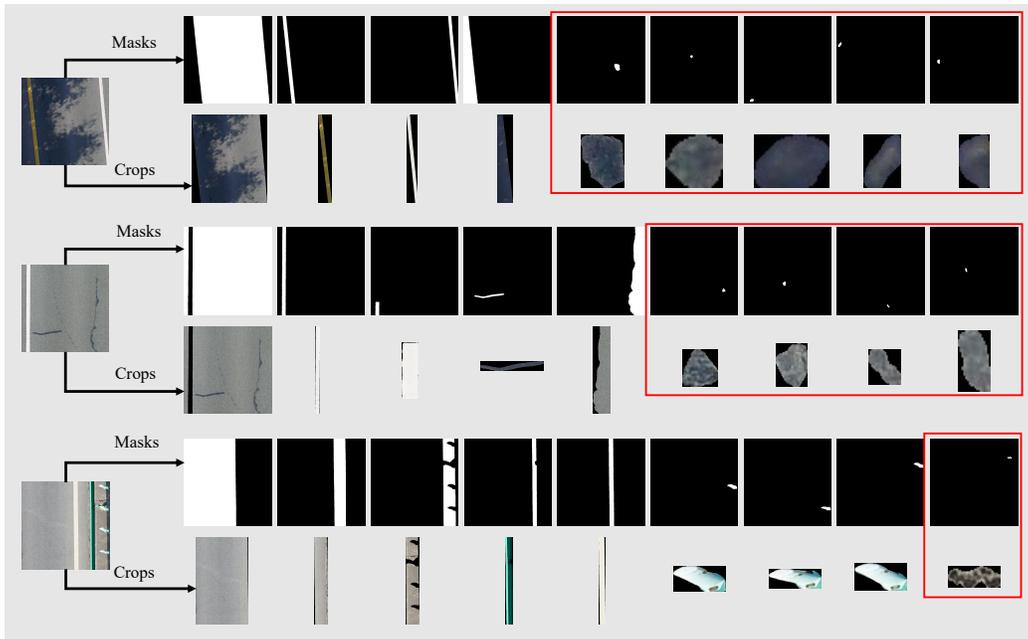}
    
    \caption{Sample masks and corresponding region crops generated by SAM with ViT-L and 16×16 point grid. Inside the red box are meaningless objects. }
    \label{fig8}
\end{figure}

Figure \ref{fig8} displays masks and corresponding region crops generated by SAM with ViT-L and a 16×16 grid on some samples from the DRID22k-dev. It is apparent that there are masks without any meaningful objects (red box in Figure \ref{fig8}), and we filtered out these masks by an area threshold (we chose 3000). This significantly reduces the computational cost for subsequent representation feature extraction and unsupervised clustering.

\subsection{Representation Learning for Mask Region}
\label{subsec:representation}
We crop object regions from the meaningful masks generated by SAM in the DRID22k-dev, resize these regions to 224 × 224 resolution, and then feed them into either ResNet50 \citep{he2016deep} or DINOv2 \citep{oquab2023dinov2}. In previous work \citep{zadaianchuk2022unsupervised}, the authors used the model's last hidden state to denote the input image's representation features. However, we aim to explore further because our input is not a complete image but rather a part of it (see crops in Figure \ref{fig8}). The image encoder we use is a 12-layer ViT-B \citep{dosovitskiy2020image}, trained by the self-supervised framework DINOv2, with a total of 12 hidden states. We retain the feature vector of the CLS token in each layer. Additionally, the ResNet50 model is a 5-stage hierarchical CNN structure, and we follow the common practice for classification tasks, applying global average pooling to the output of each stage. In the next section, we will provide a detailed comparison of the clustering results for different levels of representation features.

\subsection{Clustering Masks for Pseudo-labels}
\label{subsec:clusteringmasksforpseudolabels}
In this section, we employ unsupervised clustering algorithms to cluster object regions' representation features.
Although six categories were labeled in the DRID22k-dev, we found that the human-defined categories did not correspond to the number of clusters found through clustering. We propose a flexible two-step strategy, referred to as over-clustering and cluster-merging, respectively. This approach, while preserving the clustering algorithm's ability to capture the inherent structure of the data, adjusts the clustering results to align with the actual categories.

\begin{figure}[h!]
    \centering
    
    \includegraphics[width=\textwidth]
    {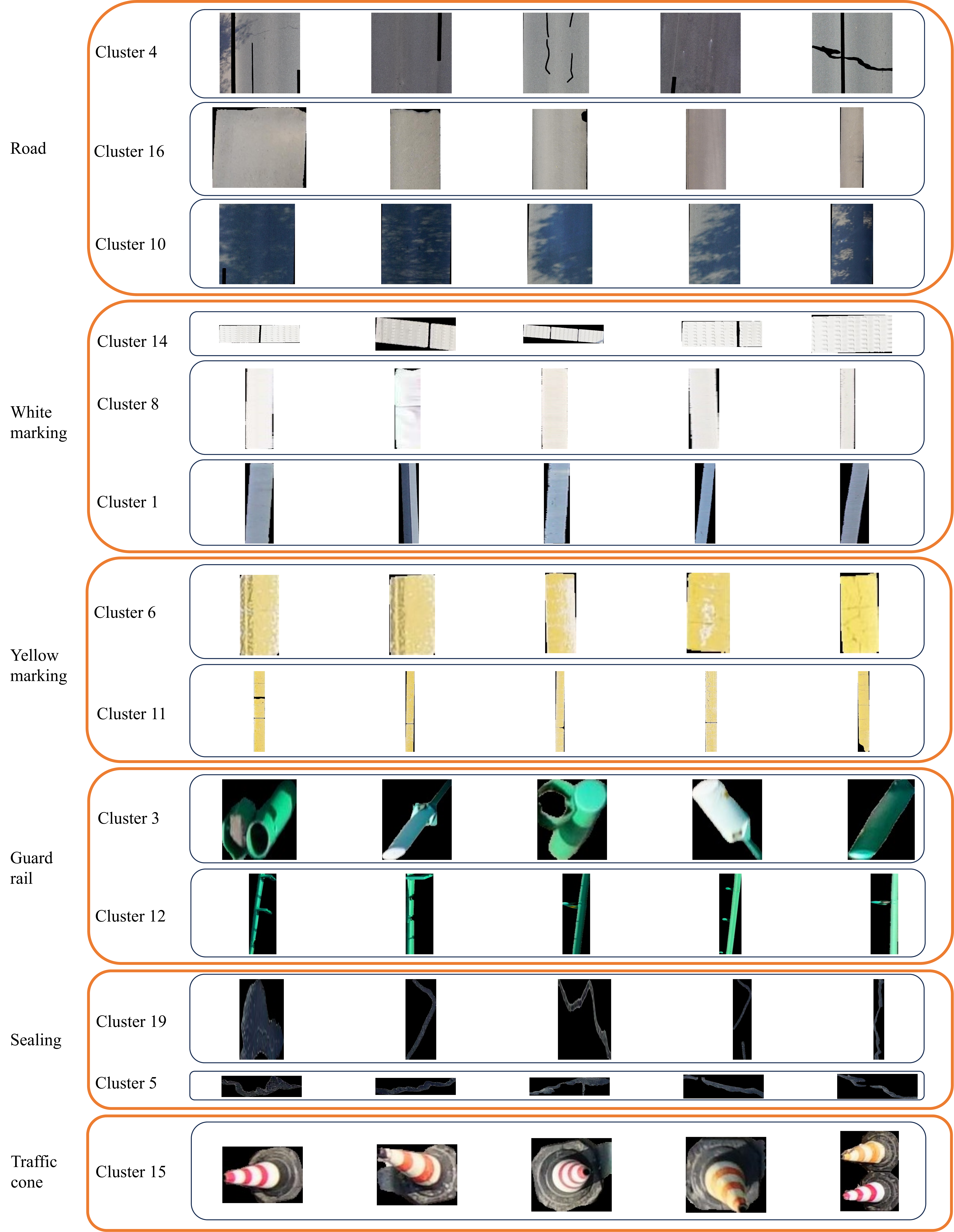}
    
    \caption{Visualization of clustering subcategories on development set. }
    \label{fig9}
\end{figure}

In the first step, we use an unsupervised clustering algorithm that does not limit the number of clusters, which can result in over-clustering, i.e., the number of subcategories is larger than the actual number of categories. The objective of this step is to ensure that we capture various variations and features within the data without overlooking any potential vital information.
In the second step, clusters containing objects of the same category are merged into a single cluster. For example, we employ spectral clustering to cluster the final hidden states of DINOv2, setting the cluster count to 20. Figure \ref{fig9} illustrates the discovered subcategories and clustering merging on the DRID22k-dev. It can be found that over-clustering reveals very fine-grained subcategories that have very clear semantic interpretability, e.g. roads in different lighting conditions (clusters 4 and 10 in Figure \ref{fig9}), different parts on guard rails (cluster 3 and 12 in Figure \ref{fig9}). Each merged cluster is assigned a specific ID. On the DRID22k-dev, a total of six IDs have been assigned, corresponding to road, white marking, yellow marking, guardrail, crack sealing and traffic cone (see Figure \ref{fig9}).

\begin{figure}[t]
    \centering
    \resizebox{\textwidth}{!}{
    \includegraphics[]{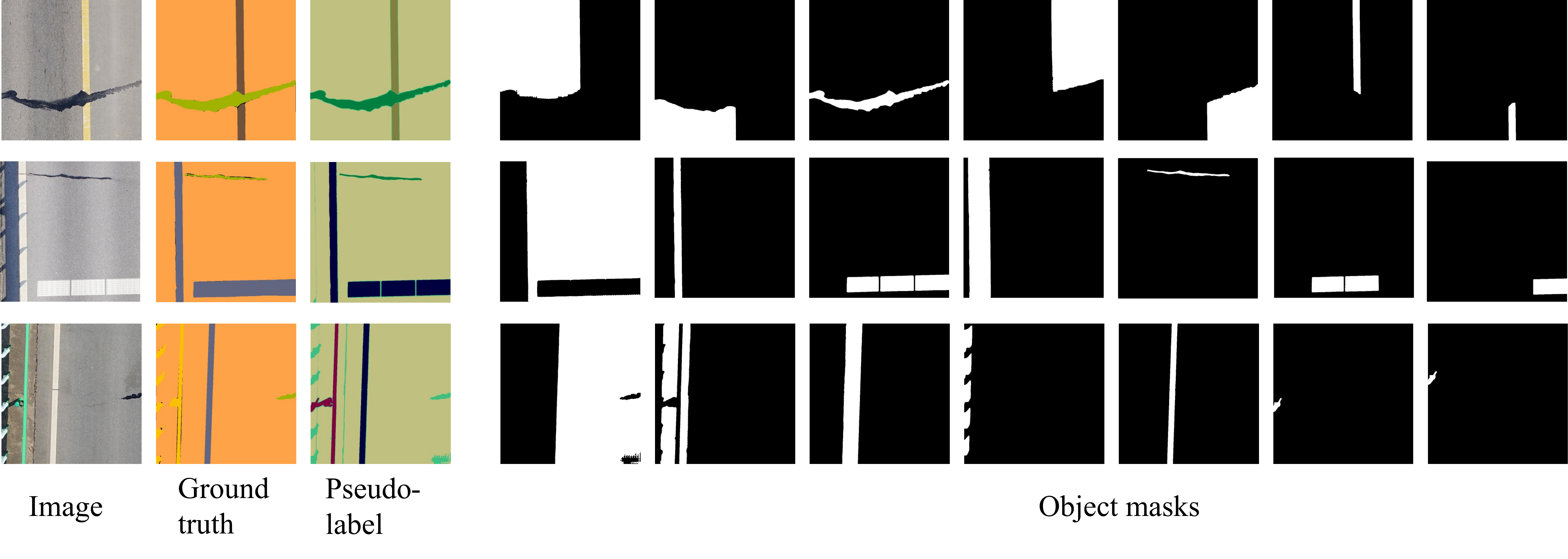}
    }
    \caption{Pseudo-label generation using masks with specified IDs (Note: not all object masks are displayed). }
    \label{fig10}
\end{figure}

Pseudo-labels are generated by combining masks with corresponding IDs after cluster merging, as shown in Figure \ref{fig10}. We evaluate the quality of these pseudo-labels on the development set by comparing them to manually annotated ground truth using intersection over union (IoU), a common evaluation metric in semantic segmentation, defined as in Equation \ref{eq1}. Additionally, we calculate pixel accuracy (PA), defined as in Equation \ref{eq2}.
In these equations, TP represents True Positives, signifying the number of correctly predicted positive pixels. TN represents True Negatives, indicating the number of correctly predicted negative pixels. FP stands for False Positives, representing the number of negative pixels erroneously predicted as positive. FN stands for False Negatives, denoting the number of positive pixels erroneously predicted as negative.

\begin{linenomath}
\begin{equation} \label{eq1}
\text{IoU}=\frac{\text{TP}}{\text{TP}+\text{FP}+\text{FN}} 
\end{equation}
\end{linenomath}

\begin{linenomath}
\begin{equation} \label{eq2}
\text{PA}=\frac{\text{TP}+\text{TN}}{\text{TP}+\text{FP}+\text{FN}+\text{TN}} 
\end{equation}
\end{linenomath}

Table \ref{tab3} and \ref{tab4} present the results of spectral clustering on all hidden states of DINOv2 and ResNet50. The overall performance of ViT-B (DINOv2) is superior to that of ResNet50, especially for challenging category like crack sealing and traffic cone. We attribute this to the different training methods employed by the two models. ResNet50 is trained in a supervised paradigm on ImageNet-1K, while ViT-B is trained using self-supervised DINOv2, along with a larger training dataset and more training tricks\citep{oquab2023dinov2}. Consequently, we focus exclusively on the ViT-B(DINOv2).

When comparing the IoU for each category, road consistently achieves the highest IoU score, even in the earlier layers of ViT-B. This is expected, as most regions in the images are roads, and their features are relatively simple and easily clustered. 

Road markings exhibit distinct color and shape attributes, thus yielding favorable clustering results.
Crack sealing has a lower IoU score due to frequent confusion with shadowed areas on the road.
Overall, the deep features of the ViT-B outperform the shallow ones. Therefore, we choose to utilize the  hidden\_12 state of the ViT-B (DINOv2).

In addition to spectral clustering, we also compared k-means, k-medoids, and agglomerative clustering on the performance of the hidden\_12 of ViT-B, as shown in Table \ref{tab5}. Overall, spectral clustering provides the best supervised signal.

\begin{table}[t]
\caption{Pseudo-label assessment based on spectral clustering for each hidden state of ViT-B (DINOv2).\label{tab3}}
\resizebox{\textwidth}{!}{
\begin{tabular}{ccccccccc}
			\toprule
\multirowcell{2}{\textbf{ViT-B} \\ \textbf{(DINOv2)}}	& \multicolumn{6}{c}{\textbf{IoU (\%)}}	& \multirow{2}{*}{\textbf{mIoU (\%)}} & \multirow{2}{*}{\textbf{PA (\%)}} \\
\cline{2-7}
 & Road & White marking & Yellow marking & Guard rail & Sealing  &Traffic cone& & \\
			\midrule
hidden\_1 &94.81 & 53.30& 53.95& 50.32& 42.21&  32.46&54.50& 92.60\\
hidden\_2 &94.33 & 65.73& 67.78& 55.33& 42.89&  39.22&60.87& 92.58\\
hidden\_3 &94.75 & 68.78& 69.13& 61.01& 43.30&  41.34&63.05& 93.23\\
hidden\_4 &95.28 & 68.91& 68.93& 61.52& 42.96&  43.59&63.53& 94.50\\
hidden\_5 &96.68 & 68.20& 75.37& 64.08& 43.00&  45.16&65.41& 95.13\\
hidden\_6 &96.07 & 70.70& 81.45& 64.20& 42.69&  56.22&68.55& 95.39\\
hidden\_7 &96.35 & 71.13& 85.86& 62.42& 42.50&  54.37&68.77& 95.24\\
hidden\_8 &96.23 & 70.56& 85.09& 64.65& 46.60&  56.72&69.97& 94.61\\
hidden\_9 &96.06 & 70.32& 83.74& 65.36& 47.62&  57.28&70.06& 95.44\\
hidden\_10 &96.38 & 72.25& 83.53& 64.74& 46.94& 60.81&70.77& 95.29\\
hidden\_11 &96.29 & 73.14& 83.34& 64.61& 47.87& 60.83&71.01& 95.23\\
hidden\_12 &96.54 & 73.00& 84.45& 64.93& 47.34& 66.03&72.04 &95.49\\

\bottomrule
\end{tabular}
  }
\end{table}

\begin{table}[t]
\caption{Pseudo-label assessment based on spectral clustering for each hidden state of ResNet50.\label{tab4}}
\resizebox{\textwidth}{!}{
		\begin{tabular}{ccccccccc}
			\toprule
\multirowcell{2}{\textbf{ResNet50} }	& \multicolumn{6}{c}{\textbf{IoU (\%)}}	& \multirow{2}{*}{\textbf{mIoU (\%)}} & \multirow{2}{*}{\textbf{PA (\%)}} \\
\cline{2-7}
 & Road & White marking & Yellow marking & Guard rail & Sealing  &Traffic cone& & \\
			\midrule
hidden\_1 &79.09 & 41.78& 23.14& 41.32& 17.13& 0 &33.74& 76.82\\
hidden\_2 &82.55 & 43.98& 34.82& 52.41& 15.50& 0 &38.21& 80.37\\
hidden\_3 &87.23 & 56.83& 32.03& 55.28& 18.84& 0 &41.70& 85.78\\
hidden\_4 &93.81 & 66.25& 38.77& 57.19& 23.41& 0 &46.57& 91.22\\
hidden\_5 &96.09 & 70.11& 52.38& 62.90& 26.22& 0 &51.28& 95.13\\
			\bottomrule
		\end{tabular}
  }
\end{table}

\begin{table}[t]
\caption{Performance of different clustering methods on hidden\_12 of ViT-B (DINOv2).\label{tab5}}
	\resizebox{\textwidth}{!}{
		\begin{tabular}{ccccccccc}
			\toprule
\multirowcell{2}{\textbf{Clustering} \\ \textbf{method} }	& \multicolumn{6}{c}{\textbf{IoU (\%)}}	& \multirow{2}{*}{\textbf{mIoU (\%)}} & \multirow{2}{*}{\textbf{PA (\%)}} \\
\cline{2-7}
 & Road & White marking & Yellow marking & Guard rail & Sealing &Traffic cone& & \\
			\midrule
spectral &96.54 & 73.00& 84.45& 64.93& 47.34& 66.03 &72.04& 95.49\\
k-means & 92.25 & 65.23& 72.57& 34.25& 30.76& 30.52 &54.26& 80.37\\
k-medoids &93.41 & 64.66& 70.49& 33.86& 32.53& 32.49 &54.57& 85.78\\
agglomerative &93.18 & 67.52& 76.22& 52.01&  35.88& 39.25 &60.67& 91.22\\

		\bottomrule
		\end{tabular}
  }
\end{table}

\subsection{Self-training with Pseudo-labels}
Although the preceding steps can indeed accomplish the road segmentation of drone imagery, the fragmented nature of these procedures remains somewhat inconvenient in practical applications, as each module requires manual tuning of hyperparameter. We condense all the knowledge into an end-to-end model by means of self-training semantic segmentation models.

In the experiment on the DRID22k-dev, we have explored the optimal framework components, including the mask generator SAM (ViT-L with 16x16 grid), the representation feature hidden\_12 of ViT-B (DINOv2), and the spectral clustering algorithm. These components are applied to DRID22k to produce pseudo-labels, which are subsequently employed in training a regular semantic segmentation model, acting as the teacher. Once the teacher model's training is complete, it can generate new pseudo-labels on DRID22k, which are then used to train the student model. We repeat this teacher-student self-training loop until the model's performance no longer improves. It is worth noting that model's performance evaluation is conducted on the DRID22k, which is annotated manually.

During self-training, we explored two different architectures of the model Deeplabv3p \citep{chen2018encoder} and SegFormer \citep{xie2021segformer} as a comparison. The specific implementations of these models are adapted from MMSegmentation \citep{mmseg2020}, and default pre-trained weights are employed. For data augmentation, we use random resize, random crop, and random flip. The random resize has a scale of [512, 1024] and cropping is performed at a resolution of 512 × 512. All training tasks are conducted on an Nvidia A5000 GPU using PyTorch. Table \ref{tab6} presents the parameters utilized in the self-training process.

\begin{table}[t] 
\caption{Self-training parameters for CORMP.\label{tab6}}
\resizebox{\textwidth}{!}{
\begin{tabular}{ccc}
\toprule
\textbf{Parameters}	& \textbf{Deeplabv3p} & \textbf{SegFormer}	\\
\midrule
Optimizer		& SGD(momentum=0.9, weight\_decay=0.0005)			&AdamW(betas=(0.9, 0.999), weight\_decay=0.01))\\
Learning rate		& 0.001	&0.00006\\
Batch size    & 	8     & 8\\
Iterations    &  80000   &80000\\
\bottomrule
\end{tabular}
}
\end{table}

\begin{figure}[t]
\resizebox{\textwidth}{!}{
\includegraphics[]{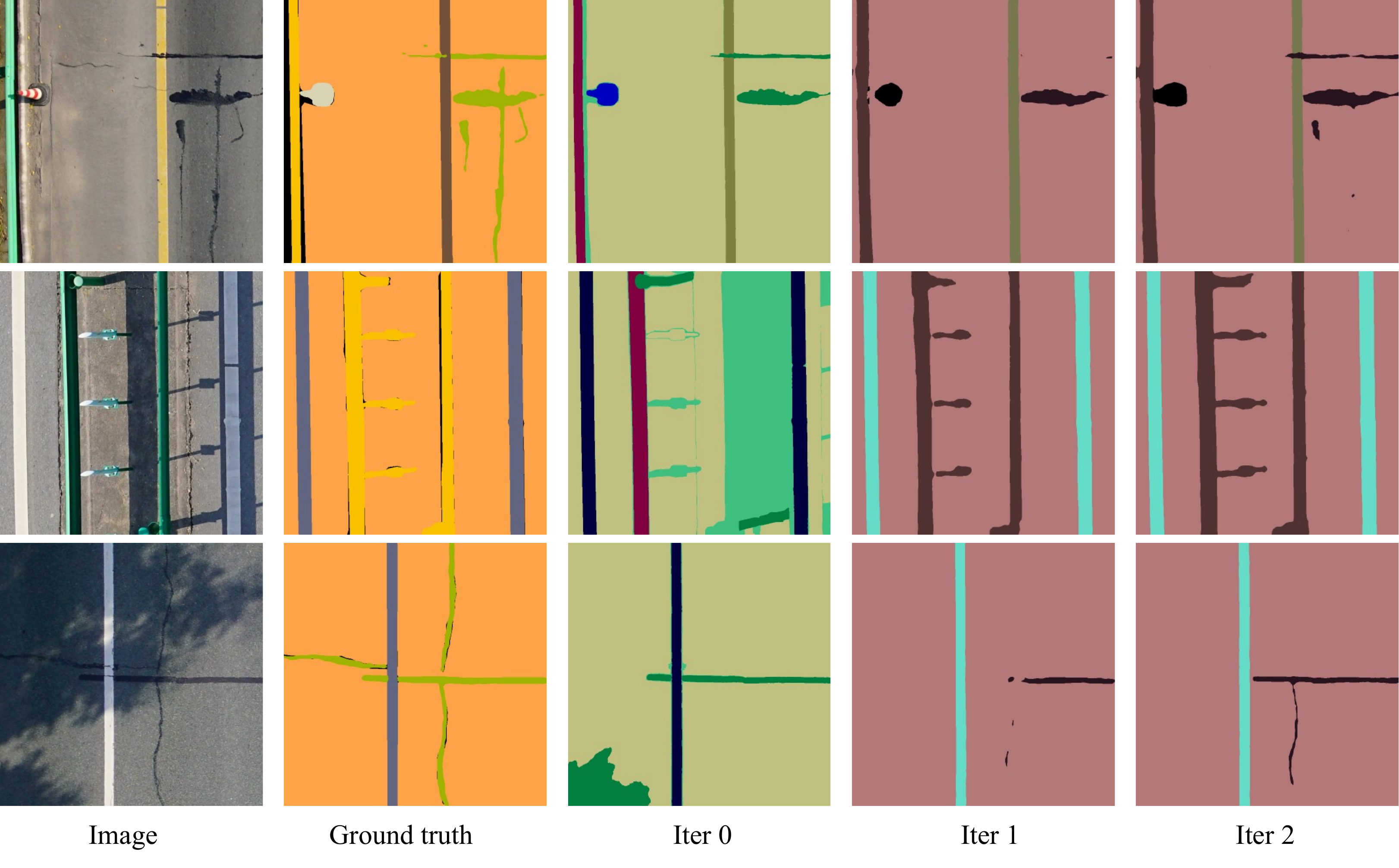}
}
\caption{Visualizations of COMRP with Deeplabv3p on DRID22k-dev. \label{fig11}}
\end{figure}

We conducted two iterations of self-training on the pseudo-labels from DRID22k, as outlined in Table \ref{tab7}. Overall, we observed that models trained with more pseudo-labels consistently outperformed those relying solely on clustering method, particularly evident in the first iteration. The second iteration still improves, but mainly focuses on the crack sealing and traffic cone. Furthermore, the models with two different architectures (CNN and ViT) get similar results, which supports our considerations for the inclusiveness of the proposed framework. To provide a more intuitive sense of the improvements brought by self-training, we visualize some samples from the DRID22k-dev in Figure \ref{fig11}.

\begin{table}[t]
\caption{Effect of number of self-training iterations and different models.\label{tab7}}
\resizebox{\textwidth}{!}{
		\begin{tabular}{ccccccccc}
			\toprule
\multirowcell{2}{\textbf{Model} }	& \multicolumn{6}{c}{\textbf{IoU (\%)}}	& \multirow{2}{*}{\textbf{mIoU (\%)}} & \multirow{2}{*}{\textbf{PA (\%)}} \\
\cline{2-7}
 & Road & White marking & Yellow marking & Guard rail & Sealing &Traffic cone& & \\
			\midrule
Clustering(iter 0) &96.54 & 73.00& 84.45& 64.93& 47.34& 66.03 &72.04& 95.49\\
Deeplabv3p (iter 1) & 99.21 & 91.40& 94.28 & 88.73& 69.20&  89.04 &88.64& 98.77\\
Deeplabv3p (iter 2) &99.29 & 91.03& 93.82& 88.50&71.03 & 91.76 &89.23& 98.82\\
SegFormer (iter 1) &99.28 & 92.51& 94.76& 88.91&  69.52& 89.99 &89.16& 98.75\\
SegFormer (iter 2) &99.32 & 92.06& 94.78& 88.55&  72.34& 92.74 &89.96& 98.90\\

		\bottomrule
		\end{tabular}
  }
\end{table}

\section{Discussion}

\begin{figure}[t]
\resizebox{\textwidth}{!}{
\includegraphics[]{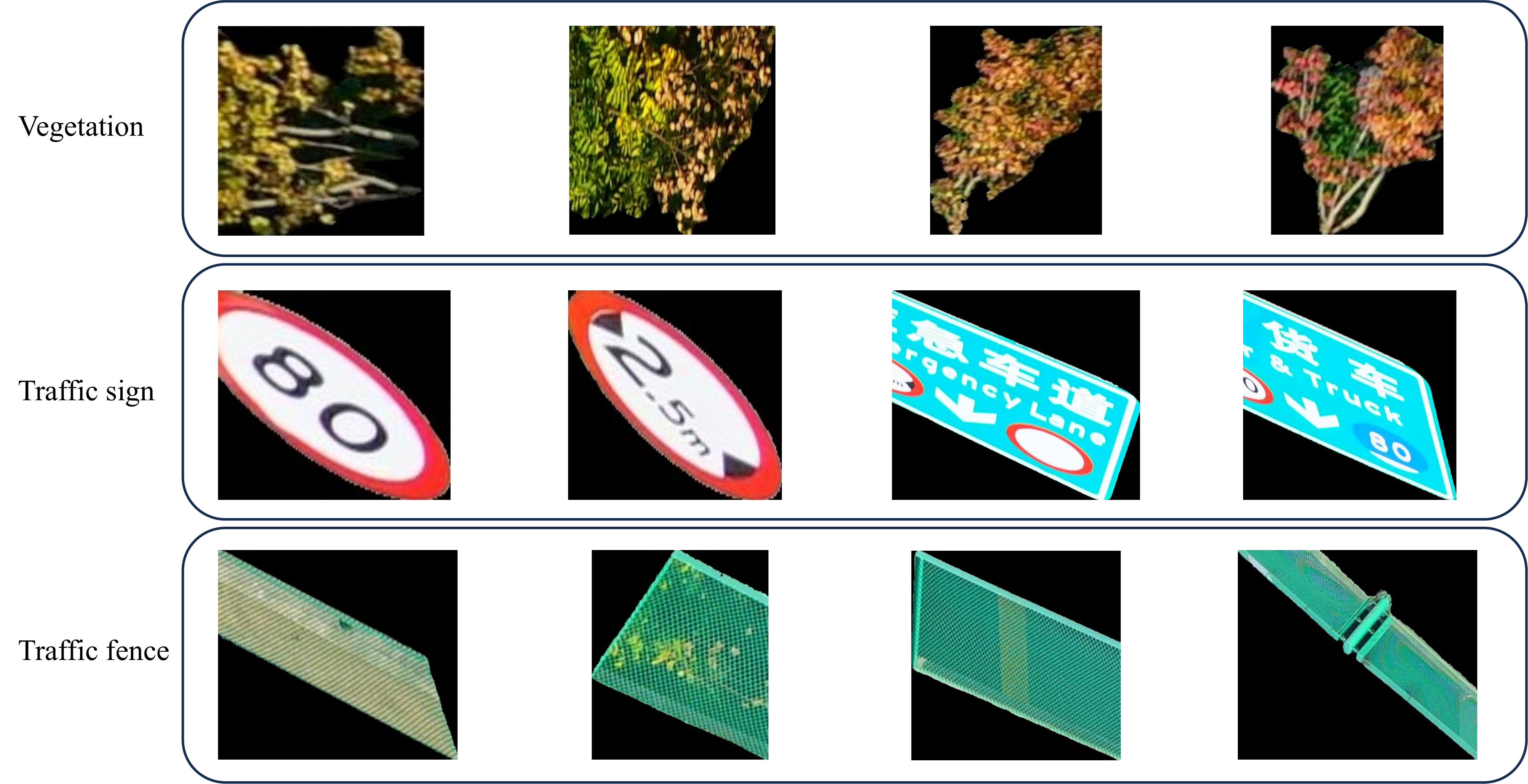}
}
\caption{Additional object categories found in DRID22k.\label{fig12}}
\end{figure}

During the analysis of the DRID22k dataset, our clustering algorithm identified several unexpected road objects such as vegetation, traffic signs, and traffic fences, as illustrated in Figure \ref{fig12}. This discovery underscores the inherent flexibility of our COMRP framework, which eliminates the need for predefined categories and allows for the natural emergence of object categories from the data itself. This capability aligns with current research trends towards object detection and segmentation in open-set scenarios \citep{liu2023grounding, Mukhoti2023openvocb}. However, most emerging methods rely on pre-training using web-scale data, which typically afford only for large research institutions like OpenAI and Meta. In contrast, our framework offers a more accessible approach to open-set semantic segmentation, with the primary computational cost being the training of a regular semantic segmentation network.

While the proposed COMRP framework demonstrates promising unsupervised road semantic segmentation, it is important to acknowledge several limitations. Firstly, we refer to our approach as unsupervised mainly because it doesn't rely on manually annotated labels, yet  there are actually some parameters that still need to be specified artificially. For instance, the threshold for the area of masks to be filtered is set to 3000, which was determined through preliminary experiments. Secondly, our method struggles to distinguish highly similar objects, such as crack sealing and shadows. Lastly, clustering merging relies on human priors, which does introduce an element of supervision to our method that may be considered less unsupervised.

\section{Conclusions}

In this work, we develope and present a novel framework for unsupervised road parsing in UAV remote sensing imagery. First, we explore the application of most current multi-modal vision-language models in high-resolution UAV images. Next, SAM, a foundation model in computer vision, is used as a mask generator to generate masks in our proposed dataset DRID22k.  Following this, a vision transformer trained with the self-supervised learning framework DINOv2 is utilized to extract representation features from the mask regions. These feature vectors are clustered by spectral clustering and assigned corresponding IDs. The masks are combined with the corresponding IDs to generate pseudo-labels that served as the foundation for the iterative self-training of a regular semantic segmentation network.

Following evaluation on the development set, our method shows highly promising performance. Specifically, without any manual annotations, we achieves an mIoU of 89.96\% across six manually annotated categories. To the best of our knowledge, we are the first to introduce unsupervised semantic segmentation into the field of road parsing in UAV imagery. Our approach demonstrate considerable potential for practical applications due to its minimal need for manual annotations and computational resources.

In the future work, the robustness and stability of the method needs to be further investigated in order to be extended to a wider range of real-world application scenarios.

\section*{CRediT authorship contribution statement}
\textbf{Zihan Ma:} Conceptualization, Validation, Writing – original draft, Writing – review \& editing. \textbf{Yongshang Li:} Conceptualization, Methodology, Resources, Writing – original draft, Writing – review \& editing, Investigation, Validation, Project administration. 
\textbf{Ronggui Ma:} Supervision, Funding acquisition, Project administration. \textbf{Chen Liang:} Writing – review \& editing, Validation. 

\section*{Funding}
This work was supported in part by the Key Research and Development Project of China under Grant 2021YFB1600104, in part by the the National Natural Science Foundation of China under Grant 52002031.

\section*{Declaration of Competing Interest}
The authors declare that they have no known competing financial interests or personal relationships that could have appeared to influence the work reported in this paper.




 \bibliographystyle{elsarticle-harv} 
 \bibliography{newref}

\begin{thebibliography}{53}
\expandafter\ifx\csname natexlab\endcsname\relax\def\natexlab#1{#1}\fi
\providecommand{\url}[1]{\texttt{#1}}
\providecommand{\href}[2]{#2}
\providecommand{\path}[1]{#1}
\providecommand{\DOIprefix}{doi:}
\providecommand{\ArXivprefix}{arXiv:}
\providecommand{\URLprefix}{URL: }
\providecommand{\Pubmedprefix}{pmid:}
\providecommand{\doi}[1]{\href{http://dx.doi.org/#1}{\path{#1}}}
\providecommand{\Pubmed}[1]{\href{pmid:#1}{\path{#1}}}
\providecommand{\bibinfo}[2]{#2}
\ifx\xfnm\relax \def\xfnm[#1]{\unskip,\space#1}\fi
\bibitem[{Astor et~al.(2023)Astor, Nabesima, Utami, Sihombing, Adli and Firdaus}]{astor2023unmanned}
\bibinfo{author}{Astor, Y.}, \bibinfo{author}{Nabesima, Y.}, \bibinfo{author}{Utami, R.}, \bibinfo{author}{Sihombing, A.V.R.}, \bibinfo{author}{Adli, M.}, \bibinfo{author}{Firdaus, M.R.}, \bibinfo{year}{2023}.
\newblock \bibinfo{title}{Unmanned aerial vehicle implementation for pavement condition survey}.
\newblock \bibinfo{journal}{Transportation Engineering} \bibinfo{volume}{12}, \bibinfo{pages}{100168}.
\newblock \DOIprefix\doi{10.1016/j.treng.2023.100168}.
\bibitem[{Bengio et~al.(2013)Bengio, Courville and Vincent}]{bengio2013representation}
\bibinfo{author}{Bengio, Y.}, \bibinfo{author}{Courville, A.}, \bibinfo{author}{Vincent, P.}, \bibinfo{year}{2013}.
\newblock \bibinfo{title}{Representation learning: A review and new perspectives}.
\newblock \bibinfo{journal}{IEEE transactions on pattern analysis and machine intelligence} \bibinfo{volume}{35}, \bibinfo{pages}{1798--1828}.
\newblock \DOIprefix\doi{10.1109/TPAMI.2013.50}.
\bibitem[{Bouraima et~al.(2023)Bouraima, Qiu, Stevi{\'c}, Marinkovi{\'c} and Deveci}]{bouraima2023integrated}
\bibinfo{author}{Bouraima, M.B.}, \bibinfo{author}{Qiu, Y.}, \bibinfo{author}{Stevi{\'c}, {\v{Z}}.}, \bibinfo{author}{Marinkovi{\'c}, D.}, \bibinfo{author}{Deveci, M.}, \bibinfo{year}{2023}.
\newblock \bibinfo{title}{Integrated intelligent decision support model for ranking regional transport infrastructure programmes based on performance assessment}.
\newblock \bibinfo{journal}{Expert Systems with Applications} \bibinfo{volume}{222}, \bibinfo{pages}{119852}.
\newblock \DOIprefix\doi{10.1016/j.eswa.2023.119852}.
\bibitem[{Brown et~al.(2020)Brown, Mann, Ryder, Subbiah, Kaplan, Dhariwal, Neelakantan, Shyam, Sastry, Askell, Agarwal, Herbert-Voss, Krueger, Henighan, Child, Ramesh, Ziegler, Wu, Winter, Hesse, Chen, Sigler, Litwin, Gray, Chess, Clark, Berner, McCandlish, Radford, Sutskever and Amodei}]{brown2020language}
\bibinfo{author}{Brown, T.B.}, \bibinfo{author}{Mann, B.}, \bibinfo{author}{Ryder, N.}, \bibinfo{author}{Subbiah, M.}, \bibinfo{author}{Kaplan, J.}, \bibinfo{author}{Dhariwal, P.}, \bibinfo{author}{Neelakantan, A.}, \bibinfo{author}{Shyam, P.}, \bibinfo{author}{Sastry, G.}, \bibinfo{author}{Askell, A.}, \bibinfo{author}{Agarwal, S.}, \bibinfo{author}{Herbert-Voss, A.}, \bibinfo{author}{Krueger, G.}, \bibinfo{author}{Henighan, T.}, \bibinfo{author}{Child, R.}, \bibinfo{author}{Ramesh, A.}, \bibinfo{author}{Ziegler, D.M.}, \bibinfo{author}{Wu, J.}, \bibinfo{author}{Winter, C.}, \bibinfo{author}{Hesse, C.}, \bibinfo{author}{Chen, M.}, \bibinfo{author}{Sigler, E.}, \bibinfo{author}{Litwin, M.}, \bibinfo{author}{Gray, S.}, \bibinfo{author}{Chess, B.}, \bibinfo{author}{Clark, J.}, \bibinfo{author}{Berner, C.}, \bibinfo{author}{McCandlish, S.}, \bibinfo{author}{Radford, A.}, \bibinfo{author}{Sutskever, I.}, \bibinfo{author}{Amodei, D.}, \bibinfo{year}{2020}.
\newblock \bibinfo{title}{Language models are few-shot learners}.
\newblock \DOIprefix\doi{10.48550/arXiv.2005.14165}, \href{http://arxiv.org/abs/2005.14165}{{\tt arXiv:2005.14165}}.
\bibitem[{Byun et~al.(2021)Byun, Shin, Moon, Kang and Choi}]{byun2021road}
\bibinfo{author}{Byun, S.}, \bibinfo{author}{Shin, I.K.}, \bibinfo{author}{Moon, J.}, \bibinfo{author}{Kang, J.}, \bibinfo{author}{Choi, S.I.}, \bibinfo{year}{2021}.
\newblock \bibinfo{title}{Road traffic monitoring from uav images using deep learning networks}.
\newblock \bibinfo{journal}{Remote Sensing} \bibinfo{volume}{13}, \bibinfo{pages}{4027}.
\newblock \DOIprefix\doi{10.3390/rs13204027}.
\bibitem[{Cao et~al.(2023)Cao, Gao, Cai, Xu and Li}]{cao2023segmentation}
\bibinfo{author}{Cao, H.}, \bibinfo{author}{Gao, Y.}, \bibinfo{author}{Cai, W.}, \bibinfo{author}{Xu, Z.}, \bibinfo{author}{Li, L.}, \bibinfo{year}{2023}.
\newblock \bibinfo{title}{Segmentation detection method for complex road cracks collected by uav based on hc-unet++}.
\newblock \bibinfo{journal}{Drones} \bibinfo{volume}{7}, \bibinfo{pages}{189}.
\newblock \DOIprefix\doi{10.3390/drones7030189}.
\bibitem[{Caron et~al.(2021)Caron, Touvron, Misra, J{\'e}gou, Mairal, Bojanowski and Joulin}]{caron2021emerging}
\bibinfo{author}{Caron, M.}, \bibinfo{author}{Touvron, H.}, \bibinfo{author}{Misra, I.}, \bibinfo{author}{J{\'e}gou, H.}, \bibinfo{author}{Mairal, J.}, \bibinfo{author}{Bojanowski, P.}, \bibinfo{author}{Joulin, A.}, \bibinfo{year}{2021}.
\newblock \bibinfo{title}{Emerging properties in self-supervised vision transformers}, in: \bibinfo{booktitle}{Proceedings of the IEEE/CVF international conference on computer vision}, pp. \bibinfo{pages}{9650--9660}.
\newblock \DOIprefix\doi{10.1109/ICCV48922.2021.00951}.
\bibitem[{Chen et~al.(2018)Chen, Zhu, Papandreou, Schroff and Adam}]{chen2018encoder}
\bibinfo{author}{Chen, L.C.}, \bibinfo{author}{Zhu, Y.}, \bibinfo{author}{Papandreou, G.}, \bibinfo{author}{Schroff, F.}, \bibinfo{author}{Adam, H.}, \bibinfo{year}{2018}.
\newblock \bibinfo{title}{Encoder-decoder with atrous separable convolution for semantic image segmentation}, in: \bibinfo{booktitle}{Proceedings of the European conference on computer vision (ECCV)}, pp. \bibinfo{pages}{801--818}.
\newblock \DOIprefix\doi{10.1007/978-3-030-01234-2_49}.
\bibitem[{Cho et~al.(2021)Cho, Mall, Bala and Hariharan}]{cho2021picie}
\bibinfo{author}{Cho, J.H.}, \bibinfo{author}{Mall, U.}, \bibinfo{author}{Bala, K.}, \bibinfo{author}{Hariharan, B.}, \bibinfo{year}{2021}.
\newblock \bibinfo{title}{Picie: Unsupervised semantic segmentation using invariance and equivariance in clustering}, in: \bibinfo{booktitle}{Proceedings of the IEEE/CVF Conference on Computer Vision and Pattern Recognition}, pp. \bibinfo{pages}{16794--16804}.
\newblock \DOIprefix\doi{10.1109/CVPR46437.2021.01652}.
\bibitem[{Chu et~al.(2024)Chu, Chen and Deng}]{chu2024cascade}
\bibinfo{author}{Chu, H.}, \bibinfo{author}{Chen, W.}, \bibinfo{author}{Deng, L.}, \bibinfo{year}{2024}.
\newblock \bibinfo{title}{Cascade operation-enhanced high-resolution representation learning for meticulous segmentation of bridge cracks}.
\newblock \bibinfo{journal}{Advanced Engineering Informatics} \bibinfo{volume}{61}, \bibinfo{pages}{102508}.
\newblock \DOIprefix\doi{10.1016/j.aei.2024.102508}.
\bibitem[{Contributors(2020)}]{mmseg2020}
\bibinfo{author}{Contributors, M.}, \bibinfo{year}{2020}.
\newblock \bibinfo{title}{{MMSegmentation}: Openmmlab semantic segmentation toolbox and benchmark}.
\newblock \bibinfo{howpublished}{\url{https://github.com/open-mmlab/mmsegmentation}}.
\bibitem[{Cordts et~al.(2016)Cordts, Omran, Ramos, Rehfeld, Enzweiler, Benenson, Franke, Roth and Schiele}]{cordts2016cityscapes}
\bibinfo{author}{Cordts, M.}, \bibinfo{author}{Omran, M.}, \bibinfo{author}{Ramos, S.}, \bibinfo{author}{Rehfeld, T.}, \bibinfo{author}{Enzweiler, M.}, \bibinfo{author}{Benenson, R.}, \bibinfo{author}{Franke, U.}, \bibinfo{author}{Roth, S.}, \bibinfo{author}{Schiele, B.}, \bibinfo{year}{2016}.
\newblock \bibinfo{title}{The cityscapes dataset for semantic urban scene understanding}, in: \bibinfo{booktitle}{Proceedings of the IEEE conference on computer vision and pattern recognition}, pp. \bibinfo{pages}{3213--3223}.
\newblock \DOIprefix\doi{10.1109/CVPR.2016.350}.
\bibitem[{Deng et~al.(2009)Deng, Dong, Socher, Li, Li and Fei-Fei}]{deng2009imagenet}
\bibinfo{author}{Deng, J.}, \bibinfo{author}{Dong, W.}, \bibinfo{author}{Socher, R.}, \bibinfo{author}{Li, L.J.}, \bibinfo{author}{Li, K.}, \bibinfo{author}{Fei-Fei, L.}, \bibinfo{year}{2009}.
\newblock \bibinfo{title}{Imagenet: A large-scale hierarchical image database}, in: \bibinfo{booktitle}{2009 IEEE conference on computer vision and pattern recognition}, \bibinfo{organization}{Ieee}. pp. \bibinfo{pages}{248--255}.
\newblock \DOIprefix\doi{10.1109/CVPR.2009.5206848}.
\bibitem[{Devlin et~al.(2018)Devlin, Chang, Lee and Toutanova}]{devlin2018bert}
\bibinfo{author}{Devlin, J.}, \bibinfo{author}{Chang, M.W.}, \bibinfo{author}{Lee, K.}, \bibinfo{author}{Toutanova, K.}, \bibinfo{year}{2018}.
\newblock \bibinfo{title}{Bert: Pre-training of deep bidirectional transformers for language understanding}.
\newblock \bibinfo{journal}{arXiv preprint arXiv:1810.04805} \DOIprefix\doi{10.48550/arXiv.1810.04805}.
\bibitem[{Dosovitskiy et~al.(2020)Dosovitskiy, Beyer, Kolesnikov, Weissenborn, Zhai, Unterthiner, Dehghani, Minderer, Heigold, Gelly et~al.}]{dosovitskiy2020image}
\bibinfo{author}{Dosovitskiy, A.}, \bibinfo{author}{Beyer, L.}, \bibinfo{author}{Kolesnikov, A.}, \bibinfo{author}{Weissenborn, D.}, \bibinfo{author}{Zhai, X.}, \bibinfo{author}{Unterthiner, T.}, \bibinfo{author}{Dehghani, M.}, \bibinfo{author}{Minderer, M.}, \bibinfo{author}{Heigold, G.}, \bibinfo{author}{Gelly, S.}, et~al., \bibinfo{year}{2020}.
\newblock \bibinfo{title}{An image is worth 16x16 words: Transformers for image recognition at scale}.
\newblock \bibinfo{journal}{arXiv preprint arXiv:2010.11929} \DOIprefix\doi{10.48550/arXiv.2010.11929}.
\bibitem[{Du et~al.(2022)Du, Shen, Wang, Fei, Li, Wu, Zhao, Fu and Liu}]{du2022learning}
\bibinfo{author}{Du, Y.}, \bibinfo{author}{Shen, Y.}, \bibinfo{author}{Wang, H.}, \bibinfo{author}{Fei, J.}, \bibinfo{author}{Li, W.}, \bibinfo{author}{Wu, L.}, \bibinfo{author}{Zhao, R.}, \bibinfo{author}{Fu, Z.}, \bibinfo{author}{Liu, Q.}, \bibinfo{year}{2022}.
\newblock \bibinfo{title}{Learning from future: A novel self-training framework for semantic segmentation}.
\newblock \bibinfo{journal}{Advances in Neural Information Processing Systems} \bibinfo{volume}{35}, \bibinfo{pages}{4749--4761}.
\newblock \DOIprefix\doi{10.48550/arXiv.2209.06993}.
\bibitem[{Everingham et~al.()Everingham, Van~Gool, Williams, Winn and Zisserman}]{pascal-voc-2012}
\bibinfo{author}{Everingham, M.}, \bibinfo{author}{Van~Gool, L.}, \bibinfo{author}{Williams, C.K.I.}, \bibinfo{author}{Winn, J.}, \bibinfo{author}{Zisserman, A.}, .
\newblock \bibinfo{title}{The {PASCAL} {V}isual {O}bject {C}lasses {C}hallenge 2012 {(VOC2012)} {R}esults}.
\newblock \bibinfo{howpublished}{http://www.pascal-network.org/challenges/VOC/voc2012/workshop/index.html}.
\bibitem[{Flah et~al.(2021)Flah, Nunez, Ben~Chaabene and Nehdi}]{flah2021machine}
\bibinfo{author}{Flah, M.}, \bibinfo{author}{Nunez, I.}, \bibinfo{author}{Ben~Chaabene, W.}, \bibinfo{author}{Nehdi, M.L.}, \bibinfo{year}{2021}.
\newblock \bibinfo{title}{Machine learning algorithms in civil structural health monitoring: A systematic review}.
\newblock \bibinfo{journal}{Archives of computational methods in engineering} \bibinfo{volume}{28}, \bibinfo{pages}{2621--2643}.
\newblock \DOIprefix\doi{10.1007/s11831-020-09471-9}.
\bibitem[{Gao et~al.(2023)Gao, Cao, Cai and Zhou}]{gao2023pixel}
\bibinfo{author}{Gao, Y.}, \bibinfo{author}{Cao, H.}, \bibinfo{author}{Cai, W.}, \bibinfo{author}{Zhou, G.}, \bibinfo{year}{2023}.
\newblock \bibinfo{title}{Pixel-level road crack detection in uav remote sensing images based on ard-unet}.
\newblock \bibinfo{journal}{Measurement} \bibinfo{volume}{219}, \bibinfo{pages}{113252}.
\newblock \DOIprefix\doi{10.1016/j.measurement.2023.113252}.
\bibitem[{Hamilton et~al.(2022)Hamilton, Zhang, Hariharan, Snavely and Freeman}]{hamilton2022unsupervised}
\bibinfo{author}{Hamilton, M.}, \bibinfo{author}{Zhang, Z.}, \bibinfo{author}{Hariharan, B.}, \bibinfo{author}{Snavely, N.}, \bibinfo{author}{Freeman, W.T.}, \bibinfo{year}{2022}.
\newblock \bibinfo{title}{Unsupervised semantic segmentation by distilling feature correspondences}.
\newblock \bibinfo{journal}{arXiv preprint arXiv:2203.08414} \DOIprefix\doi{10.48550/arXiv.2203.08414}.
\bibitem[{Hamzenejadi and Mohseni(2023)}]{HAMZENEJADI2023120845}
\bibinfo{author}{Hamzenejadi, M.H.}, \bibinfo{author}{Mohseni, H.}, \bibinfo{year}{2023}.
\newblock \bibinfo{title}{Fine-tuned yolov5 for real-time vehicle detection in uav imagery: Architectural improvements and performance boost}.
\newblock \bibinfo{journal}{Expert Systems with Applications} \bibinfo{volume}{231}, \bibinfo{pages}{120845}.
\newblock \DOIprefix\doi{10.1016/j.eswa.2023.120845}.
\bibitem[{Hao et~al.(2022)Hao, Liu, Chen, Han, Peng, Tang, Chen, Wu, Chen and Lai}]{hao2022eiseg}
\bibinfo{author}{Hao, Y.}, \bibinfo{author}{Liu, Y.}, \bibinfo{author}{Chen, Y.}, \bibinfo{author}{Han, L.}, \bibinfo{author}{Peng, J.}, \bibinfo{author}{Tang, S.}, \bibinfo{author}{Chen, G.}, \bibinfo{author}{Wu, Z.}, \bibinfo{author}{Chen, Z.}, \bibinfo{author}{Lai, B.}, \bibinfo{year}{2022}.
\newblock \bibinfo{title}{Eiseg: An efficient interactive segmentation annotation tool based on paddlepaddle}.
\newblock \bibinfo{journal}{arXiv preprint arXiv:2210.08788} \DOIprefix\doi{10.48550/arXiv.2210.08788}.
\bibitem[{He et~al.(2016)He, Zhang, Ren and Sun}]{he2016deep}
\bibinfo{author}{He, K.}, \bibinfo{author}{Zhang, X.}, \bibinfo{author}{Ren, S.}, \bibinfo{author}{Sun, J.}, \bibinfo{year}{2016}.
\newblock \bibinfo{title}{Deep residual learning for image recognition}, in: \bibinfo{booktitle}{Proceedings of the IEEE conference on computer vision and pattern recognition}, pp. \bibinfo{pages}{770--778}.
\newblock \DOIprefix\doi{10.48550/arXiv.1512.03385}.
\bibitem[{Hu and Assaad(2023)}]{HU2023120897}
\bibinfo{author}{Hu, X.}, \bibinfo{author}{Assaad, R.H.}, \bibinfo{year}{2023}.
\newblock \bibinfo{title}{The use of unmanned ground vehicles (mobile robots) and unmanned aerial vehicles (drones) in the civil infrastructure asset management sector: Applications, robotic platforms, sensors, and algorithms}.
\newblock \bibinfo{journal}{Expert Systems with Applications} \bibinfo{volume}{232}, \bibinfo{pages}{120897}.
\newblock \DOIprefix\doi{10.1016/j.eswa.2023.120897}.
\bibitem[{Iftikhar et~al.(2023)Iftikhar, Asim, Zhang, Muthanna, Chen, El-Affendi, Sedik and Abd El-Latif}]{iftikhar2023target}
\bibinfo{author}{Iftikhar, S.}, \bibinfo{author}{Asim, M.}, \bibinfo{author}{Zhang, Z.}, \bibinfo{author}{Muthanna, A.}, \bibinfo{author}{Chen, J.}, \bibinfo{author}{El-Affendi, M.}, \bibinfo{author}{Sedik, A.}, \bibinfo{author}{Abd El-Latif, A.A.}, \bibinfo{year}{2023}.
\newblock \bibinfo{title}{Target detection and recognition for traffic congestion in smart cities using deep learning-enabled uavs: A review and analysis}.
\newblock \bibinfo{journal}{Applied Sciences} \bibinfo{volume}{13}, \bibinfo{pages}{3995}.
\newblock \DOIprefix\doi{10.3390/app13063995}.
\bibitem[{Jeong et~al.(2022)Jeong, Seo and Wacker}]{JEONG2022116791}
\bibinfo{author}{Jeong, E.}, \bibinfo{author}{Seo, J.}, \bibinfo{author}{Wacker, J.P.}, \bibinfo{year}{2022}.
\newblock \bibinfo{title}{Uav-aided bridge inspection protocol through machine learning with improved visibility images}.
\newblock \bibinfo{journal}{Expert Systems with Applications} \bibinfo{volume}{197}, \bibinfo{pages}{116791}.
\newblock \DOIprefix\doi{10.1016/j.eswa.2022.116791}.
\bibitem[{Kirillov et~al.(2023)Kirillov, Mintun, Ravi, Mao, Rolland, Gustafson, Xiao, Whitehead, Berg, Lo et~al.}]{kirillov2023segment}
\bibinfo{author}{Kirillov, A.}, \bibinfo{author}{Mintun, E.}, \bibinfo{author}{Ravi, N.}, \bibinfo{author}{Mao, H.}, \bibinfo{author}{Rolland, C.}, \bibinfo{author}{Gustafson, L.}, \bibinfo{author}{Xiao, T.}, \bibinfo{author}{Whitehead, S.}, \bibinfo{author}{Berg, A.C.}, \bibinfo{author}{Lo, W.Y.}, et~al., \bibinfo{year}{2023}.
\newblock \bibinfo{title}{Segment anything}.
\newblock \bibinfo{journal}{arXiv preprint arXiv:2304.02643} \DOIprefix\doi{10.48550/arXiv.2304.02643}.
\bibitem[{Krizhevsky et~al.(2012)Krizhevsky, Sutskever and Hinton}]{krizhevsky2012imagenet}
\bibinfo{author}{Krizhevsky, A.}, \bibinfo{author}{Sutskever, I.}, \bibinfo{author}{Hinton, G.E.}, \bibinfo{year}{2012}.
\newblock \bibinfo{title}{Imagenet classification with deep convolutional neural networks}.
\newblock \bibinfo{journal}{Advances in neural information processing systems} \bibinfo{volume}{25}.
\newblock \DOIprefix\doi{10.1145/3065386}.
\bibitem[{Lin et~al.(2014)Lin, Maire, Belongie, Hays, Perona, Ramanan, Doll{\'a}r and Zitnick}]{lin2014microsoft}
\bibinfo{author}{Lin, T.Y.}, \bibinfo{author}{Maire, M.}, \bibinfo{author}{Belongie, S.}, \bibinfo{author}{Hays, J.}, \bibinfo{author}{Perona, P.}, \bibinfo{author}{Ramanan, D.}, \bibinfo{author}{Doll{\'a}r, P.}, \bibinfo{author}{Zitnick, C.L.}, \bibinfo{year}{2014}.
\newblock \bibinfo{title}{Microsoft coco: Common objects in context}, in: \bibinfo{booktitle}{Computer Vision--ECCV 2014: 13th European Conference, Zurich, Switzerland, September 6-12, 2014, Proceedings, Part V 13}, \bibinfo{organization}{Springer}. pp. \bibinfo{pages}{740--755}.
\newblock \DOIprefix\doi{10.1007/978-3-319-10602-1_48}.
\bibitem[{Liu et~al.(2023)Liu, Zeng, Ren, Li, Zhang, Yang, Li, Yang, Su, Zhu et~al.}]{liu2023grounding}
\bibinfo{author}{Liu, S.}, \bibinfo{author}{Zeng, Z.}, \bibinfo{author}{Ren, T.}, \bibinfo{author}{Li, F.}, \bibinfo{author}{Zhang, H.}, \bibinfo{author}{Yang, J.}, \bibinfo{author}{Li, C.}, \bibinfo{author}{Yang, J.}, \bibinfo{author}{Su, H.}, \bibinfo{author}{Zhu, J.}, et~al., \bibinfo{year}{2023}.
\newblock \bibinfo{title}{Grounding dino: Marrying dino with grounded pre-training for open-set object detection}.
\newblock \bibinfo{journal}{arXiv preprint arXiv:2303.05499} \DOIprefix\doi{10.48550/arXiv.2303.05499}.
\bibitem[{Liu et~al.(2021)Liu, Lin, Cao, Hu, Wei, Zhang, Lin and Guo}]{liu2021swin}
\bibinfo{author}{Liu, Z.}, \bibinfo{author}{Lin, Y.}, \bibinfo{author}{Cao, Y.}, \bibinfo{author}{Hu, H.}, \bibinfo{author}{Wei, Y.}, \bibinfo{author}{Zhang, Z.}, \bibinfo{author}{Lin, S.}, \bibinfo{author}{Guo, B.}, \bibinfo{year}{2021}.
\newblock \bibinfo{title}{Swin transformer: Hierarchical vision transformer using shifted windows}, in: \bibinfo{booktitle}{Proceedings of the IEEE/CVF international conference on computer vision}, pp. \bibinfo{pages}{10012--10022}.
\newblock \DOIprefix\doi{10.1109/ICCV48922.2021.00986}.
\bibitem[{Melas-Kyriazi et~al.(2022)Melas-Kyriazi, Rupprecht, Laina and Vedaldi}]{melas2022deep}
\bibinfo{author}{Melas-Kyriazi, L.}, \bibinfo{author}{Rupprecht, C.}, \bibinfo{author}{Laina, I.}, \bibinfo{author}{Vedaldi, A.}, \bibinfo{year}{2022}.
\newblock \bibinfo{title}{Deep spectral methods: A surprisingly strong baseline for unsupervised semantic segmentation and localization}, in: \bibinfo{booktitle}{Proceedings of the IEEE/CVF Conference on Computer Vision and Pattern Recognition}, pp. \bibinfo{pages}{8364--8375}.
\newblock \DOIprefix\doi{10.1109/CVPR52688.2022.00818}.
\bibitem[{Mukhoti et~al.(2023)Mukhoti, Lin, Poursaeed, Wang, Shah, Torr and Lim}]{Mukhoti2023openvocb}
\bibinfo{author}{Mukhoti, J.}, \bibinfo{author}{Lin, T.Y.}, \bibinfo{author}{Poursaeed, O.}, \bibinfo{author}{Wang, R.}, \bibinfo{author}{Shah, A.}, \bibinfo{author}{Torr, P.H.}, \bibinfo{author}{Lim, S.N.}, \bibinfo{year}{2023}.
\newblock \bibinfo{title}{Open vocabulary semantic segmentation with patch aligned contrastive learning}, in: \bibinfo{booktitle}{2023 IEEE/CVF Conference on Computer Vision and Pattern Recognition (CVPR)}, pp. \bibinfo{pages}{19413--19423}.
\newblock \DOIprefix\doi{10.1109/CVPR52729.2023.01860}.
\bibitem[{Nguyen et~al.(2021)Nguyen, Perry, Bone, Le and Nguyen}]{nguyen2021two}
\bibinfo{author}{Nguyen, N.H.T.}, \bibinfo{author}{Perry, S.}, \bibinfo{author}{Bone, D.}, \bibinfo{author}{Le, H.T.}, \bibinfo{author}{Nguyen, T.T.}, \bibinfo{year}{2021}.
\newblock \bibinfo{title}{Two-stage convolutional neural network for road crack detection and segmentation}.
\newblock \bibinfo{journal}{Expert Systems with Applications} \bibinfo{volume}{186}, \bibinfo{pages}{115718}.
\newblock \DOIprefix\doi{10.1016/j.eswa.2021.115718}.
\bibitem[{Oquab et~al.(2023)Oquab, Darcet, Moutakanni, Vo, Szafraniec, Khalidov, Fernandez, Haziza, Massa, El-Nouby et~al.}]{oquab2023dinov2}
\bibinfo{author}{Oquab, M.}, \bibinfo{author}{Darcet, T.}, \bibinfo{author}{Moutakanni, T.}, \bibinfo{author}{Vo, H.}, \bibinfo{author}{Szafraniec, M.}, \bibinfo{author}{Khalidov, V.}, \bibinfo{author}{Fernandez, P.}, \bibinfo{author}{Haziza, D.}, \bibinfo{author}{Massa, F.}, \bibinfo{author}{El-Nouby, A.}, et~al., \bibinfo{year}{2023}.
\newblock \bibinfo{title}{Dinov2: Learning robust visual features without supervision}.
\newblock \bibinfo{journal}{arXiv preprint arXiv:2304.07193} \DOIprefix\doi{10.48550/arXiv.2304.07193}.
\bibitem[{Outay et~al.(2020)Outay, Mengash and Adnan}]{outay2020applications}
\bibinfo{author}{Outay, F.}, \bibinfo{author}{Mengash, H.A.}, \bibinfo{author}{Adnan, M.}, \bibinfo{year}{2020}.
\newblock \bibinfo{title}{Applications of unmanned aerial vehicle (uav) in road safety, traffic and highway infrastructure management: Recent advances and challenges}.
\newblock \bibinfo{journal}{Transportation research part A: policy and practice} \bibinfo{volume}{141}, \bibinfo{pages}{116--129}.
\newblock \DOIprefix\doi{10.1016/j.tra.2020.09.018}.
\bibitem[{Qiu et~al.(2022)Qiu, Huang and Tang}]{qiu2022asff}
\bibinfo{author}{Qiu, M.}, \bibinfo{author}{Huang, L.}, \bibinfo{author}{Tang, B.H.}, \bibinfo{year}{2022}.
\newblock \bibinfo{title}{Asff-yolov5: Multielement detection method for road traffic in uav images based on multiscale feature fusion}.
\newblock \bibinfo{journal}{Remote Sensing} \bibinfo{volume}{14}, \bibinfo{pages}{3498}.
\newblock \DOIprefix\doi{10.3390/rs14143498}.
\bibitem[{Radford et~al.(2021)Radford, Kim, Hallacy, Ramesh, Goh, Agarwal, Sastry, Askell, Mishkin, Clark et~al.}]{radford2021learning}
\bibinfo{author}{Radford, A.}, \bibinfo{author}{Kim, J.W.}, \bibinfo{author}{Hallacy, C.}, \bibinfo{author}{Ramesh, A.}, \bibinfo{author}{Goh, G.}, \bibinfo{author}{Agarwal, S.}, \bibinfo{author}{Sastry, G.}, \bibinfo{author}{Askell, A.}, \bibinfo{author}{Mishkin, P.}, \bibinfo{author}{Clark, J.}, et~al., \bibinfo{year}{2021}.
\newblock \bibinfo{title}{Learning transferable visual models from natural language supervision}, in: \bibinfo{booktitle}{International conference on machine learning}, \bibinfo{organization}{PMLR}. pp. \bibinfo{pages}{8748--8763}.
\newblock \DOIprefix\doi{10.48550/arXiv.2103.00020}.
\bibitem[{Senthilnath et~al.(2020)Senthilnath, Varia, Dokania, Anand and Benediktsson}]{senthilnath2020deep}
\bibinfo{author}{Senthilnath, J.}, \bibinfo{author}{Varia, N.}, \bibinfo{author}{Dokania, A.}, \bibinfo{author}{Anand, G.}, \bibinfo{author}{Benediktsson, J.A.}, \bibinfo{year}{2020}.
\newblock \bibinfo{title}{Deep tec: Deep transfer learning with ensemble classifier for road extraction from uav imagery}.
\newblock \bibinfo{journal}{Remote Sensing} \bibinfo{volume}{12}, \bibinfo{pages}{245}.
\newblock \DOIprefix\doi{10.3390/rs12020245}.
\bibitem[{Silva et~al.(2023)Silva, Leithardt, Batista, Gonz{\'a}lez and Santana}]{silva2023automated}
\bibinfo{author}{Silva, L.A.}, \bibinfo{author}{Leithardt, V.R.Q.}, \bibinfo{author}{Batista, V.F.L.}, \bibinfo{author}{Gonz{\'a}lez, G.V.}, \bibinfo{author}{Santana, J.F.D.P.}, \bibinfo{year}{2023}.
\newblock \bibinfo{title}{Automated road damage detection using uav images and deep learning techniques}.
\newblock \bibinfo{journal}{IEEE Access} \DOIprefix\doi{10.1109/ACCESS.2023.3287770}.
\bibitem[{Spencer~Jr et~al.(2019)Spencer~Jr, Hoskere and Narazaki}]{spencer2019advances}
\bibinfo{author}{Spencer~Jr, B.F.}, \bibinfo{author}{Hoskere, V.}, \bibinfo{author}{Narazaki, Y.}, \bibinfo{year}{2019}.
\newblock \bibinfo{title}{Advances in computer vision-based civil infrastructure inspection and monitoring}.
\newblock \bibinfo{journal}{Engineering} \bibinfo{volume}{5}, \bibinfo{pages}{199--222}.
\newblock \DOIprefix\doi{10.1016/j.eng.2018.11.030}.
\bibitem[{Van~Gansbeke et~al.(2021)Van~Gansbeke, Vandenhende, Georgoulis and Van~Gool}]{van2021unsupervised}
\bibinfo{author}{Van~Gansbeke, W.}, \bibinfo{author}{Vandenhende, S.}, \bibinfo{author}{Georgoulis, S.}, \bibinfo{author}{Van~Gool, L.}, \bibinfo{year}{2021}.
\newblock \bibinfo{title}{Unsupervised semantic segmentation by contrasting object mask proposals}, in: \bibinfo{booktitle}{Proceedings of the IEEE/CVF International Conference on Computer Vision}, pp. \bibinfo{pages}{10052--10062}.
\newblock \DOIprefix\doi{10.1109/ICCV48922.2021.00990}.
\bibitem[{Von~Luxburg(2007)}]{von2007tutorial}
\bibinfo{author}{Von~Luxburg, U.}, \bibinfo{year}{2007}.
\newblock \bibinfo{title}{A tutorial on spectral clustering}.
\newblock \bibinfo{journal}{Statistics and computing} \bibinfo{volume}{17}, \bibinfo{pages}{395--416}.
\newblock \DOIprefix\doi{10.1007/s11222-007-9033-z}.
\bibitem[{Wu et~al.(2023)Wu, Meng, Qin, Qian, Xu and Jia}]{wu2023uav}
\bibinfo{author}{Wu, Y.}, \bibinfo{author}{Meng, F.}, \bibinfo{author}{Qin, Y.}, \bibinfo{author}{Qian, Y.}, \bibinfo{author}{Xu, F.}, \bibinfo{author}{Jia, L.}, \bibinfo{year}{2023}.
\newblock \bibinfo{title}{Uav imagery based potential safety hazard evaluation for high-speed railroad using real-time instance segmentation}.
\newblock \bibinfo{journal}{Advanced Engineering Informatics} \bibinfo{volume}{55}, \bibinfo{pages}{101819}.
\newblock \DOIprefix\doi{10.1016/j.aei.2022.101819}.
\bibitem[{Xie et~al.(2021)Xie, Wang, Yu, Anandkumar, Alvarez and Luo}]{xie2021segformer}
\bibinfo{author}{Xie, E.}, \bibinfo{author}{Wang, W.}, \bibinfo{author}{Yu, Z.}, \bibinfo{author}{Anandkumar, A.}, \bibinfo{author}{Alvarez, J.M.}, \bibinfo{author}{Luo, P.}, \bibinfo{year}{2021}.
\newblock \bibinfo{title}{Segformer: Simple and efficient design for semantic segmentation with transformers}.
\newblock \bibinfo{journal}{Advances in Neural Information Processing Systems} \bibinfo{volume}{34}, \bibinfo{pages}{12077--12090}.
\newblock \DOIprefix\doi{10.48550/arXiv.2105.15203}.
\bibitem[{Xie et~al.(2020)Xie, Luong, Hovy and Le}]{xie2020self}
\bibinfo{author}{Xie, Q.}, \bibinfo{author}{Luong, M.T.}, \bibinfo{author}{Hovy, E.}, \bibinfo{author}{Le, Q.V.}, \bibinfo{year}{2020}.
\newblock \bibinfo{title}{Self-training with noisy student improves imagenet classification}, in: \bibinfo{booktitle}{Proceedings of the IEEE/CVF conference on computer vision and pattern recognition}, pp. \bibinfo{pages}{10687--10698}.
\newblock \DOIprefix\doi{10.1109/CVPR42600.2020.01070}.
\bibitem[{Xu et~al.(2023)Xu, Zheng, Liu, Li and Wang}]{xu2023unmanned}
\bibinfo{author}{Xu, H.}, \bibinfo{author}{Zheng, W.}, \bibinfo{author}{Liu, F.}, \bibinfo{author}{Li, P.}, \bibinfo{author}{Wang, R.}, \bibinfo{year}{2023}.
\newblock \bibinfo{title}{Unmanned aerial vehicle perspective small target recognition algorithm based on improved yolov5}.
\newblock \bibinfo{journal}{Remote Sensing} \bibinfo{volume}{15}, \bibinfo{pages}{3583}.
\newblock \DOIprefix\doi{10.3390/rs15143583}.
\bibitem[{Zadaianchuk et~al.(2022)Zadaianchuk, Kleindessner, Zhu, Locatello and Brox}]{zadaianchuk2022unsupervised}
\bibinfo{author}{Zadaianchuk, A.}, \bibinfo{author}{Kleindessner, M.}, \bibinfo{author}{Zhu, Y.}, \bibinfo{author}{Locatello, F.}, \bibinfo{author}{Brox, T.}, \bibinfo{year}{2022}.
\newblock \bibinfo{title}{Unsupervised semantic segmentation with self-supervised object-centric representations}.
\newblock \bibinfo{journal}{arXiv preprint arXiv:2207.05027} \DOIprefix\doi{10.48550/arXiv.2207.05027}.
\bibitem[{Zhang et~al.(2023)Zhang, Huang, Jin and Lu}]{zhang2023vision}
\bibinfo{author}{Zhang, J.}, \bibinfo{author}{Huang, J.}, \bibinfo{author}{Jin, S.}, \bibinfo{author}{Lu, S.}, \bibinfo{year}{2023}.
\newblock \bibinfo{title}{Vision-language models for vision tasks: A survey}.
\newblock \bibinfo{journal}{arXiv preprint arXiv:2304.00685} \DOIprefix\doi{10.48550/arXiv.2304.00685}.
\bibitem[{Zhang et~al.(2022)Zhang, Zuo, Xu, Wu, Zhu, Zhang, Wang and Tian}]{zhang2022road}
\bibinfo{author}{Zhang, Y.}, \bibinfo{author}{Zuo, Z.}, \bibinfo{author}{Xu, X.}, \bibinfo{author}{Wu, J.}, \bibinfo{author}{Zhu, J.}, \bibinfo{author}{Zhang, H.}, \bibinfo{author}{Wang, J.}, \bibinfo{author}{Tian, Y.}, \bibinfo{year}{2022}.
\newblock \bibinfo{title}{Road damage detection using uav images based on multi-level attention mechanism}.
\newblock \bibinfo{journal}{Automation in Construction} \bibinfo{volume}{144}, \bibinfo{pages}{104613}.
\newblock \DOIprefix\doi{10.1016/j.autcon.2022.104613}.
\bibitem[{Zhou et~al.(2022)Zhou, Guo, Hou and Wu}]{zhou2022review}
\bibinfo{author}{Zhou, Y.}, \bibinfo{author}{Guo, X.}, \bibinfo{author}{Hou, F.}, \bibinfo{author}{Wu, J.}, \bibinfo{year}{2022}.
\newblock \bibinfo{title}{Review of intelligent road defects detection technology}.
\newblock \bibinfo{journal}{Sustainability} \bibinfo{volume}{14}, \bibinfo{pages}{6306}.
\newblock \DOIprefix\doi{10.3390/su14106306}.
\bibitem[{Zhu et~al.(2022)Zhu, Zhu, Bu and Gao}]{zhu2022monitoring}
\bibinfo{author}{Zhu, C.}, \bibinfo{author}{Zhu, J.}, \bibinfo{author}{Bu, T.}, \bibinfo{author}{Gao, X.}, \bibinfo{year}{2022}.
\newblock \bibinfo{title}{Monitoring and identification of road construction safety factors via uav}.
\newblock \bibinfo{journal}{Sensors} \bibinfo{volume}{22}, \bibinfo{pages}{8797}.
\newblock \DOIprefix\doi{10.3390/s22228797}.
\bibitem[{Zhu et~al.(2021)Zhu, Zhang, Wu, Zhang, He, Zhang, Manmatha, Li and Smola}]{zhu2021improving}
\bibinfo{author}{Zhu, Y.}, \bibinfo{author}{Zhang, Z.}, \bibinfo{author}{Wu, C.}, \bibinfo{author}{Zhang, Z.}, \bibinfo{author}{He, T.}, \bibinfo{author}{Zhang, H.}, \bibinfo{author}{Manmatha, R.}, \bibinfo{author}{Li, M.}, \bibinfo{author}{Smola, A.J.}, \bibinfo{year}{2021}.
\newblock \bibinfo{title}{Improving semantic segmentation via efficient self-training}.
\newblock \bibinfo{journal}{IEEE transactions on pattern analysis and machine intelligence} \DOIprefix\doi{10.1109/TPAMI.2021.3138337}.

\end{thebibliography}





\end{document}